\documentclass[preprint]{article}
\usepackage{aistats2027}
\usepackage[T1]{fontenc}
\usepackage[utf8]{inputenc}
\usepackage{amssymb,amsthm,mathtools}
\usepackage{graphicx,booktabs,multirow}
\usepackage{xspace,xurl}
\usepackage{algorithm,algpseudocode}
\usepackage[hidelinks]{hyperref}
\usepackage{algorithm,algpseudocode}
\usepackage{xcolor} % <--- ADD THIS for bright text colors
\usepackage[hidelinks]{hyperref}
\usepackage[
  backend=biber,
  style=authoryear-comp,
  natbib=true,
  sorting=nyt,
  maxbibnames=9,
  giveninits=true,
  uniquename=init,
  doi=true,
  url=true,
  isbn=false,
  eprint=true
]{biblatex}
\newcommand{\LAIR}{LAIR-Net\xspace}

\newcommand{\SLAN}{SLAN\xspace}                  % shallow learned alignment network
\newcommand{\LAIRNoAlign}{LAIR-NoAlign\xspace}
\newcommand{\LAIRRandomImpulse}{LAIR-Random-Impulse\xspace}
\newcommand{\LAIRBestState}{LAIR-BestState\xspace}
\newcommand{\LAIRSLANInit}{LAIR-SLANInit\xspace}
\newcommand{\LAIROrtho}{LAIR-Ortho\xspace}
\newcommand{\LAIROrthoBMA}{LAIR-OrthoBMA\xspace}
\newcommand{\RVFL}{RVFL\xspace}
\newcommand{\dRVFL}{dRVFL\xspace}
\newcommand{\edRVFL}{edRVFL\xspace}
\newcommand{\edRVFLSC}{edRVFL-SC\xspace}
\newcommand{\RealNum}{\mathbb{R}}                     % real number set
\newcommand{\DataSet}{\mathcal{D}}                    % supervised training dataset
\newcommand{\InputMat}{\mathbf{X}}                    % input data matrix
\newcommand{\TargetVec}{\mathbf{y}}                   % regression target vector
\newcommand{\BiasVec}{\mathbf{1}}                     % all-ones bias vector
\newcommand{\AugInput}{\mathbf{A}}                    % bias-augmented input matrix
\newcommand{\DesignMat}[1]{\mathbf{D}^{(#1)}}         % readout design matrix
\newcommand{\HiddenState}[1]{\mathbf{H}^{(#1)}}       % hidden-state matrix
\newcommand{\AnchorMat}{\mathbf{S}}                   % alignment representation
\newcommand{\LayerInput}[1]{\mathbf{Z}^{(#1)}}        % LAIR layer input
\newcommand{\RandHidden}[1]{\widetilde{\mathbf{H}}^{(#1)}}  % randomized candidate
\newcommand{\LeakState}[1]{\mathbf{U}^{(#1)}}         % leaky residual state
\newcommand{\ReadoutWeight}[1]{\boldsymbol{\beta}^{(#1)}}   % ridge readout weights
\newcommand{\LayerPred}[1]{\widehat{\mathbf{y}}^{(#1)}}     % layerwise prediction
\newcommand{\FinalPred}{\widehat{\mathbf{y}}}         % final aggregated prediction
\newcommand{\NumSamples}{N}                           % number of samples
\newcommand{\NumFeatures}{d}                          % number of input features
\newcommand{\HiddenWidth}{m}                          % hidden-layer width
\newcommand{\NumLayers}{L}                            % number of hidden layers
\newcommand{\RidgeReg}{\lambda}                       % ridge regularisation
\newcommand{\AlignReg}{\lambda_{\mathrm{s}}}          % SLAN regularisation
\newcommand{\LeakCoef}{\gamma}                        % leaky residual coefficient
\newcommand{\AlignCoef}{\alpha}                       % alignment coefficient
\newcommand{\Activation}{g}                           % nonlinear activation
\newcommand{\Aggregator}{\mathcal{A}}                 % aggregation operator
\newcommand{\IdentityMat}[1]{\mathbf{I}_{#1}}         % identity matrix
\newcommand{\Norm}[1]{\left\lVert #1 \right\rVert}    % norm operator

\newcommand{\NormTwo}[1]{\Norm{#1}_{2}}

\newcommand{\InputVec}{\mathbf{x}}                    % single input vector
\newcommand{\HiddenVec}[1]{\mathbf{h}^{(#1)}}
\newcommand{\AnchorVec}{\mathbf{s}}
\newcommand{\LayerInputVec}[1]{\mathbf{z}^{(#1)}}
\newcommand{\RandHiddenVec}[1]{\widetilde{\mathbf{h}}^{(#1)}}
\newcommand{\LeakStateVec}[1]{\mathbf{u}^{(#1)}}

\newcommand{\VarOp}{\mathrm{Var}}

\newcommand{\Transpose}{^{\top}}
\newcommand{\LayerIdx}{\ell}

\newcommand{\SlanAct}{\psi}                           % SLAN activation (NOT sigma)
\newcommand{\SlanW}{\mathbf{W}^{\mathrm{s}}}
\newcommand{\SlanB}{\mathbf{b}^{\mathrm{s}}}
\newcommand{\SlanV}{\mathbf{v}^{\mathrm{s}}}
\newcommand{\SlanC}{c^{\mathrm{s}}}
\newcommand{\SlanParam}{\Theta^{\mathrm{s}}}

\newcommand{\RandWeight}[1]{\mathbf{W}^{(#1)}}        % fixed random weights
\newcommand{\RandWeightHj}[1]{\mathbf{W}^{(#1)}_{\mathrm{h}}}
\newcommand{\RandWeightXj}[1]{\mathbf{W}^{(#1)}_{\mathrm{x}}}

\newcommand{\PropFac}{\rho}                           % propagation factor
\newcommand{\LipAct}{L_{\Activation}}                 % activation Lipschitz const
\newcommand{\StateGain}{c_{1}}                        % state-block gain
\newcommand{\InputGain}{c_{\mathrm{x}}}               % input-block gain
\newcommand{\BiasInputGain}{c_{0}}
\newcommand{\RandWeightBXj}[1]{\mathbf{W}^{(#1)}_{\mathrm{0x}}}
\newcommand{\BoundInput}{B_{x}}
\newcommand{\BoundAnchor}{B_{s}}

\newcommand{\PerturbHidden}[1]{\Delta\HiddenVec{#1}}   % hidden-state difference
\newcommand{\PerturbInput}{\Delta\InputVec}            % input difference
\newcommand{\PerturbAnchor}{\Delta\AnchorVec}          % anchor difference
\newcommand{\ReadoutDim}{d_{\mathrm{r}}}
\newcommand{\NlShare}{\kappa}                         % nonlinear share of variance
\newcommand{\NoiseSd}{\sigma}                         % noise standard deviation
\newcommand{\LinSig}{\mathcal{L}}                     % linear signal component
\newcommand{\NlSig}{\mathcal{G}}                      % nonlinear signal component
\newcommand{\DgpLatent}{\mathbf{C}}                   % informative covariates
\newcommand{\NlShareHat}{\widehat{\NlShare}}          % empirical nonlinearity proxy
\newcommand{\NoiseSdHat}{\widehat{\NoiseSd}}          % empirical noise proxy

\newcommand{\MSE}{\mathrm{MSE}}

\newcommand{\NewTerm}[1]{\emph{#1}}

\newtheorem{theorem}{Theorem}[section]
\newtheorem{proposition}[theorem]{Proposition}
\newtheorem{corollary}[theorem]{Corollary}
\newtheorem{assumption}[theorem]{Assumption}

\begin{document}
\runningtitle{LAIR-Net for Tabular Regression}
\runningauthor{Goswami, Bhambu, Karmakar}
\twocolumn[
\aistatstitle{LAIR-Net: Leaky Alignment-Impulse Residual Networks for Tabular Regression\\[0.5em]
\large \href{https://lairnet.statml.in}{\textcolor{magenta}{\textbf{Project Page: lairnet.statml.in}}}}

\aistatsauthor{Rahul Goswami\textsuperscript{1, 2} \And Aryan Bhambu\textsuperscript{3} \And Bittu Karmakar\textsuperscript{1}}

\aistatsaddress{\textsuperscript{1}Indian Institute of Technology Guwahati \\
\textsuperscript{2}SAFIR, Sorbonne University Abu Dhabi \\
\textsuperscript{3}Indian Institute of Technology Bombay}
]

\begin{abstract}
%Randomized neural networks fix hidden-layer parameters at random and fit closed-form readouts,

Deep randomized models fix hidden-layer parameters through random initialization and learn only closed-form readouts,  typically adding depth by stacking random transformations without target-aware control of hidden-state evolution. We propose \LAIR, the Leaky Alignment-Impulse Residual Network, which mixes a shallow learned anchor into each hidden state through a leaky residual transition. We derive a depth-uniform bound on input-perturbation sensitivity and use controlled simulations to attribute gains over a randomized baseline to the anchor rather than recursion or added capacity. Benefits emerge when a nonlinear target structure is learnable at the available noise level and diminish for nearly linear targets or dominant noise. Across 23 benchmark datasets, \LAIR achieves the best average rank among eight randomized networks and twelve conventional models, with relative performance associated with the same nonlinear-structure and noise quantities identified in simulation.
\end{abstract}

\section{Introduction}
\label{sec:introduction}
Randomized neural networks reduce training cost by fixing randomly initialized hidden parameters and learning only the readout. Here Randomization concerned randomized networks, unlike randomized tree based algorithm\citep{giri2026moving,geurts2006extremely} as for example RVFL combines nonlinear random features with a direct input-output connection and estimates output weights by least-squares or ridge regression \citep{pao1994rvfl,igelnik1995stochastic,malik2023rvfl}. Deep RVFL stacks randomized layers, while edRVFL aggregates separate readouts across depths \citep{shi2021drvfl}. Skip-connected variants reuse earlier hidden representations through concatenation \citep{hu2023esc}, increasing effective input width without directly regulating transitions between consecutive states. Consequently, perturbation sensitivity depends on the realized random mappings and activation function.

We propose the Leaky Alignment-Impulse Residual Randomized Network (LAIR-Net) for tabular regression. A shallow supervised network supplies a target-aware alignment anchor. At each depth, a leaky residual update blends the preceding state with a new random transformation, and an alignment impulse mixes this state with the anchor. Inspired by leaky reservoir updates \citep{jaeger2007leaky}, this construction combines state persistence with supervised alignment while retaining fixed hidden width and closed-form layerwise ridge readouts. The alignment network is trained separately before constructing the randomized hierarchy.

Conditional on fixed random weights, we establish finite-depth boundedness under bounded inputs, anchors and realized weights, a Lipschitz activation, and convex mixing coefficients. An effective propagation factor below one additionally yields depth-uniform boundedness and geometric attenuation of initial-state differences for trajectories sharing the same input, anchor and weights. We also bound deviation from the anchor under bounded layerwise mismatch and establish readout uniqueness for a positive ridge parameter. These results characterize recursion stability.

Across $23$ regression datasets and $18$ competing methods, LAIR-Net achieves the lowest average MSE rank ($4.22$, compared with $6.70$ for the next-best multilayer perceptron), with all pairwise Wilcoxon comparisons significant after Holm adjustment ($p<0.05$).

The main contributions are: (i) a fixed-width randomized architecture combining leaky state transitions, target-aware alignment, and layerwise ridge readouts; (ii) a conditional stability analysis covering boundedness, contraction, anchor deviation, and readout uniqueness; and (iii) controlled perturbation experiments and predictive evaluation against $18$ baselines on $23$ regression datasets. 

The remainder of the article is organized as follows. Sections~\ref{sec:related_work}-\ref{sec:theory} present related work, the method, and theory. Sections~\ref{sec:experiments} and~\ref{sec:simulation} report benchmarks and simulations. The article ends with conclusions, limitations, and future work in Section~\ref{sec:conclusion}.

%Section~\ref{sec:conclusion} discusses conclusions, limitations, and future work.
% END INLINE: draft_intro.tex

% BEGIN INLINE: draft_related.tex
\section{Related Work}
\label{sec:related_work}

Research on deep randomized neural networks has mainly focused on increasing representational depth while retaining fixed hidden transformations and inexpensive readout estimation. Deep RVFL architectures construct successive randomized representations across multiple hidden layers \citep{shi2021drvfl}. Ensemble variants additionally fit readouts at different depths and combine their predictions. This design allows information from several levels of the randomized hierarchy to contribute to the final prediction. The main emphasis of these models is therefore on representation construction and layerwise aggregation.

A related direction introduces additional connections between non-adjacent hidden layers. Skip-connected edRVFL supplies selected earlier hidden representations to deeper layers \citep{hu2023esc}. The additional pathways preserve lower-layer information and reduce exclusive dependence on the immediately preceding state. This approach addresses feature reuse by expanding the information available to later layers. It does not directly constrain the change between consecutive hidden states. In the formulation considered in this work, retaining earlier states also enlarges the effective layer input and the associated readout representation.

Residual connections address a related problem in fully trained neural networks by adding identity pathways between successive transformations. Their purpose is to preserve information and improve optimization in deep architectures. A residual pathway, however, does not by itself determine how strongly a new transformation should replace the previous representation. The distinction is relevant for randomized networks because the hidden transformations are fixed after initialization. The state evolution is therefore governed directly by the realized random mappings.

Explicit control of state evolution has been studied in reservoir computing. Leaky-integrator updates combine a newly generated state with the preceding state through a leakage parameter \citep{jaeger2007leaky}. The leakage parameter regulates the rate of state change and provides persistence across successive updates. This mechanism controls local state evolution, but it does not provide a supervised reference derived from the prediction target.

The distinction between persistence and supervision motivates the design considered in this paper. Existing deep and skip-connected RVFL formulations determine how representations are generated and reused across depth \citep{shi2021drvfl,hu2023esc}. Leaky state updates determine how quickly a state changes \citep{jaeger2007leaky}. These mechanisms do not, in the formulations considered here, combine controlled state transition with repeated guidance from a target-aware representation.

The proposed LAIR-Net addresses this gap by separating the two roles. The leaky residual update controls the contribution of the new randomized representation relative to the preceding hidden state. The alignment impulse introduces a supervised anchor at every depth. The resulting recursion therefore regulates both the rate of hidden-state evolution and its relation to a target-aware reference. This differs from skip-connected feature reuse because the proposed mechanism modifies the hidden-state update itself rather than only enlarging the set of representations supplied to a later layer.

\section{Proposed method}
\label{sec:method}

This section presents the \LAIR architecture. The model first constructs a sequence of hidden states and then maps the representation at each depth to a prediction. The hidden-state construction consists of three components. A shallow learned alignment network provides a target-aware reference representation. A leaky residual transition controls the contribution of the newly generated randomised representation relative to the previous hidden state. An alignment impulse then combines the resulting state with the learned reference. A ridge readout is fitted at each depth, and the layerwise predictions are aggregated to obtain the final output.

Let $\DataSet={(\InputVec_i,y_i)}_{i=1}^{\NumSamples}$ denote a supervised regression dataset, where $\InputVec_i\in\RealNum^{\NumFeatures}$ and $y_i\in\RealNum$. The stacked input matrix is denoted by $\InputMat\in\RealNum^{\NumSamples\times\NumFeatures}$ and the target vector by $\TargetVec\in\RealNum^{\NumSamples}$. Following the \RVFL formulation, the readout retains direct access to the observed inputs through the bias-augmented matrix

$$
\AugInput
=
[\BiasVec,\InputMat]
\in
\RealNum^{\NumSamples\times(\NumFeatures+1)},
$$

where $\BiasVec\in\RealNum^{\NumSamples\times 1}$ denotes the all-ones vector. The hidden width is denoted by $\HiddenWidth$, and the number of randomised layers is denoted by $\NumLayers$.

\subsection{Building the hidden state}

The hidden transformations in a deep randomised network are generated independently of the target and remain fixed after initialisation \citep{shi2021drvfl}. The randomised hierarchy can therefore construct multiple nonlinear representations, but the hidden transformations themselves are not selected using $\TargetVec$. In \LAIR, a compact supervised representation is learned before the randomised hierarchy is constructed. This representation is not used as the final prediction. Instead, it provides a common target-aware reference for the hidden states generated at different depths.

The reference representation is obtained from a \NewTerm{shallow learned alignment network} (\SLAN). The hidden activations of the \SLAN define the \NewTerm{alignment anchor}
\begin{equation}
\AnchorMat
=
\SlanAct\left(
\InputMat\SlanW
+
\BiasVec(\SlanB)\Transpose
\right),
\qquad
\AnchorMat
\in
\RealNum^{\NumSamples\times\HiddenWidth},
\label{eq:anchor}
\end{equation}
where $\SlanW\in\RealNum^{\NumFeatures\times\HiddenWidth}$ denotes the hidden-weight matrix, $\SlanB\in\RealNum^{\HiddenWidth}$ denotes the hidden bias vector, and $\SlanAct(\cdot)$ denotes the activation function. The width of $\AnchorMat$ matches the width of the randomized hidden states. The anchor can therefore be combined directly with each hidden state without requiring an additional projection.

The trainable parameters of the \SLAN are collected in $\SlanParam=\{\SlanW,\SlanB,\SlanV,\SlanC\}$. For an input $\InputVec$, the scalar output of the shallow module is

$$
f_{\SlanParam}(\InputVec)
=
(\SlanV)\Transpose
\SlanAct\left(
(\SlanW)\Transpose\InputVec+\SlanB
\right)
+
\SlanC.
$$

The \SLAN parameters are estimated by minimizing the regularised squared-error objective
\begin{equation}
\min_{\SlanParam}
\frac{1}{2\NumSamples}
\sum_{i=1}^{\NumSamples}
\left(
f_{\SlanParam}(\InputVec_i)-y_i
\right)^2
+
\frac{\AlignReg}{2\NumSamples}
\NormTwo{\SlanParam}^{2},
\label{eq:slan-objective}
\end{equation}
where $\AlignReg>0$ controls the regularisation of the shallow module. We optimize \eqref{eq:slan-objective} using L-BFGS \citep{liu1989limited}. After optimization, the scalar output of the \SLAN is not used in the final \LAIR prediction. Only the hidden representation $\AnchorMat$ in \eqref{eq:anchor} is retained. The same anchor is supplied to every randomized layer and therefore provides a fixed supervised reference throughout the depth-wise recursion.

After the anchor has been obtained, the randomized hierarchy is initialized as

$$
\HiddenState{0}
=
\mathbf{0}
\in
\RealNum^{\NumSamples\times\HiddenWidth}.
$$

For layer $\LayerIdx\in{1,\ldots,\NumLayers}$, the effective layer input contains the bias, the original input variables, and the hidden state propagated from the preceding layer. It is defined as
\begin{equation}
\LayerInput{\LayerIdx}
=
[
\BiasVec,
\HiddenState{\LayerIdx-1}
\InputMat
]
\in
\RealNum^{\NumSamples\times(1+\HiddenWidth+\NumFeatures)}.
\label{eq:layer-input}
\end{equation}

Including $\InputMat$ at every depth ensures that each randomized transformation retains direct access to the observed covariates rather than depending only on the propagated hidden state. Only the immediately preceding hidden state is included. The dimension of $\LayerInput{\LayerIdx}$ therefore remains fixed as the depth increases.

The randomized response at depth $\LayerIdx$ is computed as
\begin{equation}
\RandHidden{\LayerIdx}
=
\Activation\left(
\LayerInput{\LayerIdx}
\RandWeight{\LayerIdx}
\right),
\label{eq:rand-response}
\end{equation}
where $\RandWeight{\LayerIdx}\in\RealNum^{(1+\HiddenWidth+\NumFeatures)\times\HiddenWidth}$ is sampled once and then held fixed. The activation function $\Activation(\cdot)$ is applied elementwise. Equation~\eqref{eq:rand-response} therefore produces a new nonlinear randomized representation at every depth without optimizing the corresponding hidden weights.

Directly replacing $\HiddenState{\LayerIdx-1}$ with $\RandHidden{\LayerIdx}$ would make the state transition depend entirely on the newly generated random transformation. \LAIR instead retains part of the previous hidden state through a leaky update. Leaky state updates have been used in reservoir computing to regulate the contribution of newly generated and previously retained states \citep{jaeger2007leaky,lukosevicius2009reservoir}. Following this principle, we define the \NewTerm{leaky residual transition} as
\begin{equation}
\LeakState{\LayerIdx}
=
\LeakCoef\RandHidden{\LayerIdx}
+
(1-\LeakCoef)\HiddenState{\LayerIdx-1},
\qquad
0\leq\LeakCoef\leq 1.
\label{eq:leak}
\end{equation}

The leakage coefficient $\LeakCoef$ controls the relative contribution of the current randomized response and the previous hidden state. A value close to one gives greater weight to $\RandHidden{\LayerIdx}$. A value close to zero retains a larger contribution from $\HiddenState{\LayerIdx-1}$. The leaky transition, therefore, controls the balance between the introduction of new randomized information and the retention of the preceding state.

Because \eqref{eq:leak} is a convex combination, the leaky state satisfies

$$
\NormTwo{\LeakState{\LayerIdx}}
\leq
\LeakCoef
\NormTwo{\RandHidden{\LayerIdx}}
+
(1-\LeakCoef)
\NormTwo{\HiddenState{\LayerIdx-1}},
$$

and consequently

$$
\NormTwo{\LeakState{\LayerIdx}}
\leq
\max
\left\{
\NormTwo{\RandHidden{\LayerIdx}},
\NormTwo{\HiddenState{\LayerIdx-1}}
\right\}.
$$

The leaky transition regulates the movement between consecutive hidden states, but it does not provide a supervised reference for the resulting representation. The alignment anchor learned in \eqref{eq:anchor} is introduced for this purpose. At every depth, the leaky state is combined with the fixed target-aware representation through the \NewTerm{alignment impulse}
\begin{equation}
\HiddenState{\LayerIdx}
=
(1-\AlignCoef)\LeakState{\LayerIdx}
+
\AlignCoef\AnchorMat,
\qquad
0\leq\AlignCoef\leq 1.
\label{eq:align}
\end{equation}

The alignment coefficient $\AlignCoef$ controls the contribution of the target-aware anchor. When $\AlignCoef=0$, the hidden-state recursion reduces to the leaky randomised hierarchy in \eqref{eq:leak}. When $\AlignCoef=1$, every hidden state is equal to $\AnchorMat$, and the layer-specific randomised contribution disappears. Intermediate values retain contributions from both the randomised hierarchy and the supervised anchor.

The leakage coefficient and the alignment coefficient therefore regulate different parts of the hidden-state construction. The coefficient $\LeakCoef$ controls the balance between the new randomised response and the state inherited from the preceding depth. The coefficient $\AlignCoef$ controls the subsequent balance between the leaky state and the target-aware anchor. Combining \eqref{eq:leak} and \eqref{eq:align} gives

$$
\HiddenState{\LayerIdx}
=
(1-\AlignCoef)
\left[
\LeakCoef\,\RandHidden{\LayerIdx}
+
(1-\LeakCoef)\,\HiddenState{\LayerIdx-1}
\right]
+
\AlignCoef\,\AnchorMat.
$$

The hidden state at each depth therefore combines information from the current randomised transformation, the representation propagated from the preceding layer, and the fixed target-aware anchor. The same anchor is retained across depth, whereas each layer uses its own fixed random transformation.

\begin{figure}[t]
\centering
\includegraphics[width=\columnwidth]{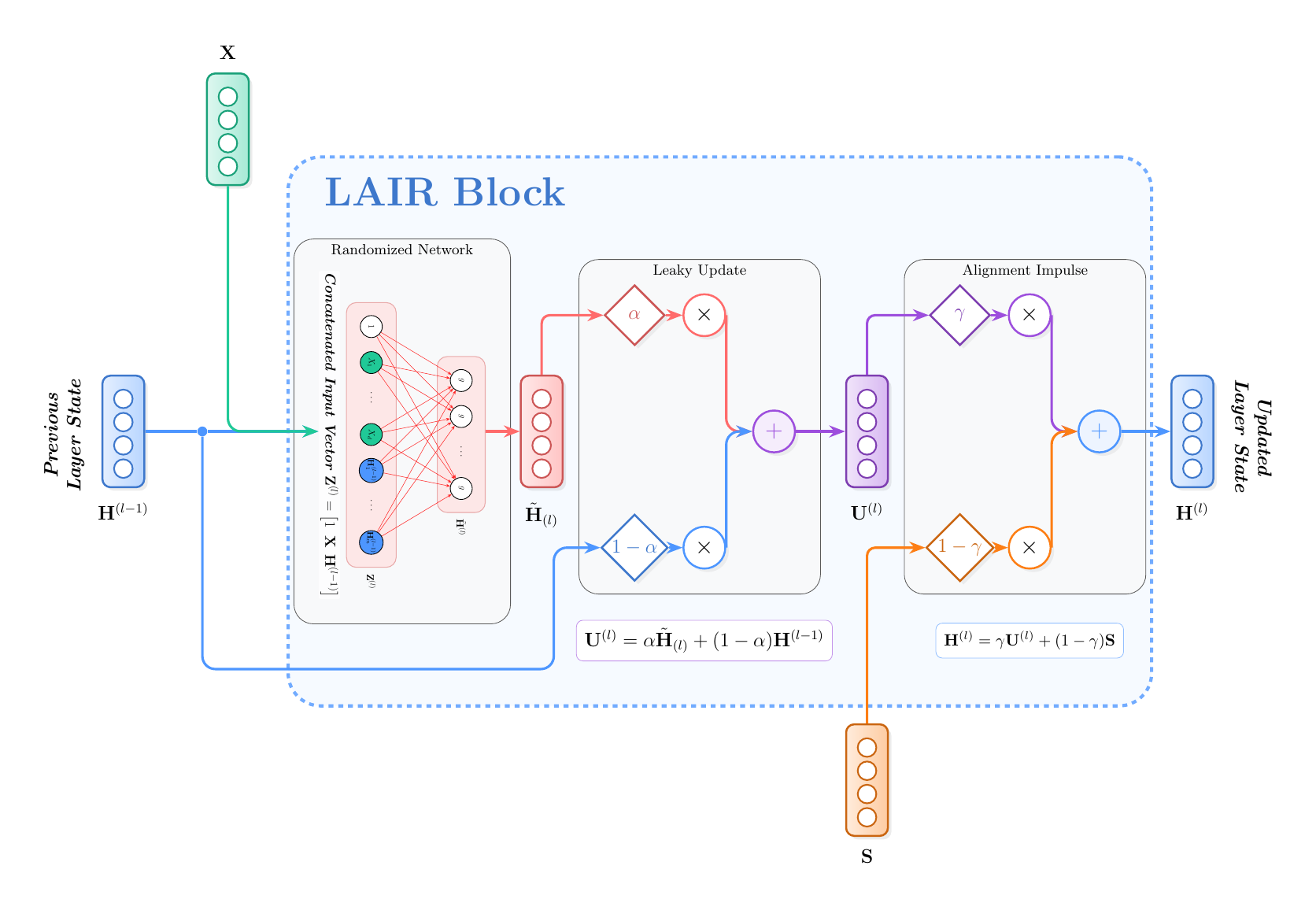}
\caption{A single \LAIR block. The layer input contains the bias, the observed inputs, and the previous hidden state. A fixed random transformation produces $\RandHidden{\LayerIdx}$. The leaky transition combines this representation with $\HiddenState{\LayerIdx-1}$ using $\LeakCoef$. The alignment impulse then combines the resulting state with $\AnchorMat$ using $\AlignCoef$.}
\label{fig:lair_block}
\end{figure}

Figure~\ref{fig:lair_block} summarises the hidden-state construction at one depth. The same sequence of random transformation, leaky transition, and alignment impulse is repeated for $\LayerIdx=1,\ldots,\NumLayers$. The resulting hidden states share the same target-aware reference while retaining layer-specific randomised components.

\subsection{From hidden states to a prediction}

The hidden-state recursion produces a distinct representation at every depth. Using only the final hidden state would discard the representations obtained at earlier depths. \LAIR therefore follows the ensemble deep \RVFL principle and fits a separate readout at each layer \citep{shi2021drvfl}. The resulting predictions provide multiple depth-dependent views of the same regression problem.

At depth $\LayerIdx$, the readout design matrix is formed by concatenating the bias-augmented input with the current hidden state
\begin{equation}
\DesignMat{\LayerIdx}
=
[
\AugInput,
\HiddenState{\LayerIdx}
]
\in
\RealNum^{\NumSamples\times(\NumFeatures+1+\HiddenWidth)}.
\label{eq:design}
\end{equation}

The original covariates therefore remain directly available to every readout. Only the current hidden state is appended to $\AugInput$, so the width of $\DesignMat{\LayerIdx}$ remains equal to $\NumFeatures+1+\HiddenWidth$ for every $\LayerIdx$.

The output weights at each depth are estimated using ridge regression \citep{hoerl1970ridge}. The corresponding closed-form solution is
\begin{equation}
\ReadoutWeight{\LayerIdx}
=
\left(
(\DesignMat{\LayerIdx})\Transpose
\DesignMat{\LayerIdx}
+
\RidgeReg
\IdentityMat{\NumFeatures+1+\HiddenWidth}
\right)^{-1}
(\DesignMat{\LayerIdx})\Transpose
\TargetVec,
\label{eq:ridge}
\end{equation}
where $\RidgeReg>0$ denotes the ridge regularisation parameter. The corresponding layerwise prediction is

$$
\LayerPred{\LayerIdx}
=
\DesignMat{\LayerIdx}
\ReadoutWeight{\LayerIdx}.
$$

The random hidden weights $\RandWeight{\LayerIdx}$ are not optimised. After the \SLAN has been fitted, the output weights associated with each randomised layer are therefore obtained through the closed-form ridge solution in \eqref{eq:ridge}.

The $\NumLayers$ readouts produce predictions from different depths of the randomised hierarchy. The final prediction is obtained as
\begin{equation}
\FinalPred
=
\Aggregator\left(
\LayerPred{1},
\ldots,
\LayerPred{\NumLayers}
\right),
\label{eq:aggregate}
\end{equation}
where $\Aggregator(\cdot)$ denotes elementwise aggregation. In the experiments, we use the elementwise median. The median limits the influence of a minority of layerwise predictions that differ substantially from the remaining outputs. The corresponding prediction-level robustness property is discussed in Section~\ref{sec:theory}.

\begin{figure}[t]
\centering
\includegraphics[width=0.82\columnwidth]{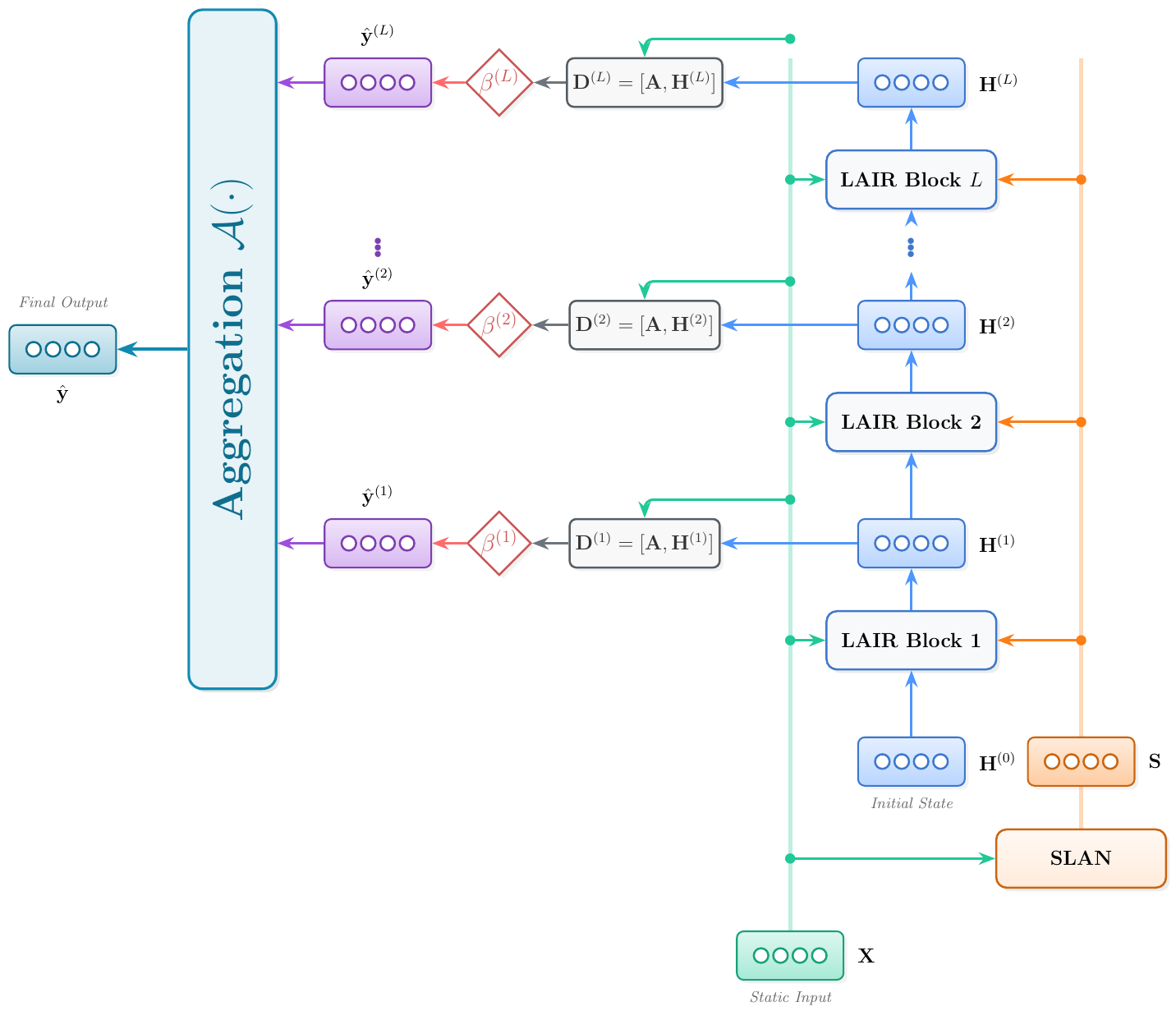}
\caption{The \LAIR architecture. The \SLAN produces the anchor $\AnchorMat$, which is supplied to every randomized layer. Each \LAIR block applies a fixed random transformation, a leaky transition, and an alignment impulse. A ridge readout is fitted at every depth, and the resulting layerwise predictions are combined using $\Aggregator$.}
\label{fig:architecture}
\end{figure}

Figure~\ref{fig:architecture} shows the complete \LAIR architecture. The \SLAN is fitted once to obtain the alignment anchor. The random hidden transformations are then sampled and fixed. Each hidden state is constructed using the randomized response in \eqref{eq:rand-response}, the leaky transition in \eqref{eq:leak}, and the alignment impulse in \eqref{eq:align}. The resulting hidden state is combined with the bias-augmented input through \eqref{eq:design}, and the corresponding readout weights are obtained using \eqref{eq:ridge}. Finally, the layerwise predictions are combined using \eqref{eq:aggregate}.

\section{Theoretical analysis}
\label{sec:theory}

We theoretically analyze the propagation of the hidden state under leaky transitions and an alignment impulse. The results are established conditionally on a fixed realization of the random weights, consistent with the prediction setting in which these weights are sampled once and subsequently remain unchanged. Under this conditioning, the analysis is deterministic, and the corresponding proofs are provided in Appendix \ref{app:proofs}.

The construction of Section~\ref{sec:method} acts on each sample independently until the readout is fitted, so it is enough to analyze the recursion for a single input $\InputVec\in\RealNum^{\NumFeatures}$ with anchor
$\AnchorVec\in\RealNum^{\HiddenWidth}$. Writing
$\LayerInputVec{\LayerIdx}=[1,\HiddenVec{\LayerIdx-1},\InputVec]$, the recursion is

\[
  \begin{aligned}
  \RandHiddenVec{\LayerIdx}
  &=\Activation\!\left((\RandWeight{\LayerIdx})\Transpose
  \LayerInputVec{\LayerIdx}\right),\\
  \LeakStateVec{\LayerIdx}
  &=\LeakCoef\RandHiddenVec{\LayerIdx}
  +(1-\LeakCoef)\HiddenVec{\LayerIdx-1},
  \end{aligned}
\]
\[
  \HiddenVec{\LayerIdx}
  =(1-\AlignCoef)\LeakStateVec{\LayerIdx}+\AlignCoef\AnchorVec ,
  \qquad \HiddenVec{0}=\mathbf{0}.
\]

\begin{assumption}[Bounded input and anchor]
\label{as:bounded}
There exist $\BoundInput>0$ and $\BoundAnchor>0$ with
$\NormTwo{\InputVec}\leq\BoundInput$ and
$\NormTwo{\AnchorVec}\leq\BoundAnchor$ for every sample considered.
\end{assumption}

\begin{assumption}[Lipschitz activation]
\label{as:lipschitz}
The activation $\Activation$ is applied elementwise and is Lipschitz with
constant $\LipAct>0$.
\end{assumption}

\begin{assumption}[Convex mixing and regularised readout]
\label{as:convex}
The coefficients satisfy $0\leq\LeakCoef\leq 1$ and
$0\leq\AlignCoef\leq 1$, and the ridge parameter satisfies $\RidgeReg>0$.
\end{assumption}

Assumption~\ref{as:bounded} describes the operating regime of the experiments
after preprocessing, where inputs are standardised and the anchor is a bounded
activation of a fitted network. It is required only for the bounded-state result
of Theorem~\ref{thm:bounded}; the input-perturbation result of
Theorem~\ref{thm:input-perturbation} bounds a difference and does not use it.
Assumption~\ref{as:lipschitz} covers the rectified linear and hyperbolic tangent
activations used here, both with $\LipAct=1$. Assumption~\ref{as:convex}
restates the design of Equations~\eqref{eq:leak} and~\eqref{eq:align}.

Partition the random matrix of layer $j$ by the blocks of
$\LayerInputVec{j}$, so that
$\RandWeightHj{j}\in\RealNum^{\HiddenWidth\times\HiddenWidth}$ acts on the
hidden state and
$\RandWeightXj{j}\in\RealNum^{\NumFeatures\times\HiddenWidth}$ acts on the
input. Because the results below iterate a single factor across depth, the
constants are taken uniformly over the realised stack,

\begin{equation}
  \StateGain=\LipAct\max_{1\leq j\leq\NumLayers}\NormTwo{\RandWeightHj{j}},
  \qquad
  \InputGain=\LipAct\max_{1\leq j\leq\NumLayers}\NormTwo{\RandWeightXj{j}},
  \label{eq:constants}
\end{equation}

and the effective propagation factor is

\begin{equation}
  \PropFac=(1-\AlignCoef)\bigl(\LeakCoef\StateGain+1-\LeakCoef\bigr).
  \label{eq:propfac}
\end{equation}

The bounded-state result additionally needs the bias row, because the layer
input carries a constant entry that does not cancel when a norm rather than a
difference is bounded. Writing $\RandWeightBXj{j}$ for the rows of
$\RandWeight{j}$ acting on $[1,\InputVec]$, we set

\begin{equation}
  \BiasInputGain=\LipAct\max_{1\leq j\leq\NumLayers}
  \NormTwo{\RandWeightBXj{j}}\sqrt{1+\BoundInput^{2}} .
  \label{eq:biasgain}
\end{equation}

All three maxima are finite for any finite depth and any realisation, so
\eqref{eq:constants} and \eqref{eq:biasgain} are well defined. The state gain uses the norm of the state
block alone, because a difference between two hidden states enters the next
layer only through that block. Using the norm of the whole random matrix would
also be valid but looser, and the looseness matters because the conclusions
below are conditional on $\PropFac<1$.

\begin{figure}[t]
\centering
\includegraphics[width=0.78\columnwidth]{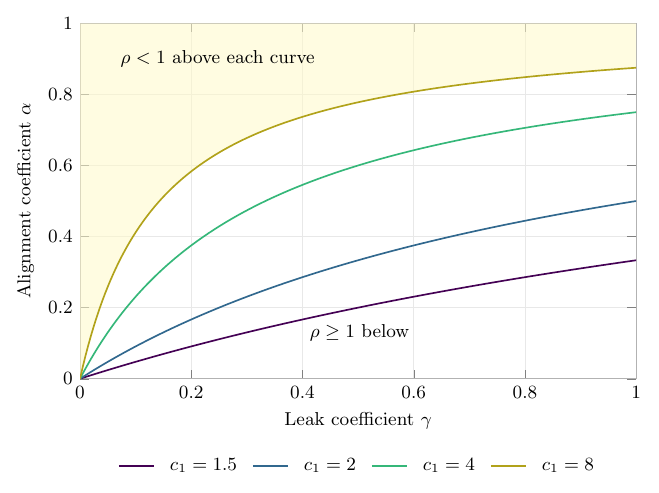}
\caption{The region in which the propagation factor of \eqref{eq:propfac} is
below one, for four values of the state gain $\StateGain$. The condition holds
above each curve. Shading marks the region for $\StateGain=8$, the most
demanding case shown. The curves are the identity
$\PropFac=(1-\AlignCoef)(1+\LeakCoef(\StateGain-1))$ rather than a measured
quantity, so the figure describes the scope of
Theorem~\ref{thm:input-perturbation} and not the behaviour of any fitted model.}
\label{fig:propfac}
\end{figure}

That condition is a statement about the coefficients and the realised state
gain, so it can be read off directly. Rearranging \eqref{eq:propfac},
$\PropFac<1$ holds exactly when
$\AlignCoef>\LeakCoef(\StateGain-1)/\bigl(1+\LeakCoef(\StateGain-1)\bigr)$.
Figure~\ref{fig:propfac} draws that boundary. The alignment coefficient is what
buys the condition, and the larger the state gain, the more of it is required.

The question the architecture is designed around is how a small change in the
input is transmitted through depth. The following statement answers it directly,
and it is the quantity measured in Section~\ref{sec:simulation}.

%%%%%%%%%%%%%%%%%%%%%%

\begin{theorem}[Input-perturbation sensitivity]
\label{thm:input-perturbation}
Let Assumptions ~\ref{as:lipschitz}-\ref{as:convex} hold. Consider two inputs $x$ and $\bar{x}$ with corresponding anchors $s$ and $\bar{s}$, propagated through the same realization of the random weights from the common initial state \(h^{(0)}=\bar{h}^{(0)}=0.\) Define \(\Delta h^{(\ell)} = h^{(\ell)}-\bar{h}^{(\ell)}\), \(\Delta x=x-\bar{x}\), \(\Delta s=s-\bar{s}\). Let \(\rho = (1-\alpha)(\gamma c_1+1-\gamma)\) and \(\delta = (1-\alpha)\gamma c_x\|\Delta x\|_2 + \alpha\|\Delta s\|_2\). Then, for every layer $1\leq\ell\leq L$,
\[
\|\Delta h^{(\ell)}\|_2
\leq
\rho\|\Delta h^{(\ell-1)}\|_2+\delta.
\]
Consequently,
\[
\|\Delta h^{(\ell)}\|_2
\leq
\begin{cases}
\displaystyle
\delta\frac{1-\rho^\ell}{1-\rho},
& \rho\neq1,\\[8pt]
\ell\delta,
& \rho=1.
\end{cases}
\]
In particular, if $\rho<1$, then
\[
\|\Delta h^{(\ell)}\|_2
\leq
\frac{\delta}{1-\rho},
\qquad
1\leq\ell\leq L,
\]
so the bound is uniform over the network depth. If, in addition, the alignment mapping $x\mapsto s(x)$ is $L_S$-Lipschitz, then \(\delta \leq \left[(1-\alpha)\gamma c_x+\alpha L_S\right]\|\Delta x\|_2\).
\end{theorem}

Theorem~\ref{thm:input-perturbation} separates the two ways a perturbation
reaches the hidden state. The term in $\InputGain$ is the direct path through
the residual input of Equation~\eqref{eq:layer-input}, and the term in
$\AlignCoef$ is the path through the anchor. Neither is amplified across depth
once $\PropFac<1$, because the recursion then contracts what it has already
accumulated faster than the new contribution arrives.

The same bound covers the no-alignment, no-leak setting. Putting
$\AlignCoef=0$ and $\LeakCoef=1$ gives $\PropFac=\StateGain$, and when this
factor exceeds one the right-hand side grows with $\LayerIdx$ instead of
remaining flat. This describes the scope of the bound rather than the behaviour
of any particular trajectory, and Section~\ref{sec:simulation} measures the
realised sensitivity in both settings.

\begin{corollary}[Hidden-state incremental stability]
\label{cor:state-perturbation}
Consider two trajectories propagated through the same realization of the
random weights with the same input $x$ and the same anchor $s$, but with
possibly different initial states $h^{(0)}$ and $\bar{h}^{(0)}$. Then, for
every layer $1\leq\ell\leq L$,
\[
\NormTwo{\PerturbHidden{\LayerIdx}}
\leq
\PropFac^{\LayerIdx}
\NormTwo{\PerturbHidden{0}}.
\]
In particular, if $\rho<1$, the hidden-state perturbation decays
geometrically with depth.
\end{corollary}

Corollary~\ref{cor:state-perturbation} considers the homogeneous state-perturbation recursion. It is stated for completeness, as it isolates the contraction mechanism in the hidden-state dynamics. This comparison is not encountered in the implemented algorithm, where the initial hidden state is fixed at the zero vector for every input. The inhomogeneous perturbation bound in Theorem~\ref{thm:input-perturbation} is the result directly relevant to the input-sensitivity experiments reported here.

The same recursion answers a second question. A bound on a difference says nothing about the size of the state itself, and an unbounded state would make the readout ill-conditioned, however well perturbations behaved.

%%%%%%%%%%%%%%%%

\begin{theorem}[Bounded hidden state]
\label{thm:bounded}
Let Assumptions ~\ref{as:bounded}-\ref{as:convex} hold and suppose that $g(0)=0$. For a fixed realization of the random weights, let \(b=(1-\alpha)\gamma c_0+\alpha B_s\). Then, for every layer $1\leq\ell\leq L$,
\[
\|h^{(\ell)}\|_2 \leq \rho\|h^{(\ell-1)}\|_2+b.
\]
Since $h^{(0)}=0$, it follows that
\[
\|h^{(\ell)}\|_2
\leq
\begin{cases}
\displaystyle
b\frac{1-\rho^\ell}{1-\rho},
& \rho\neq1,\\[8pt]
\ell b,
& \rho=1.
\end{cases}
\]
In particular, if $\rho<1$, then
\[
\|h^{(\ell)}\|_2 \leq \frac{b}{1-\rho}, \qquad 1\leq\ell\leq L,
\]
So the hidden state is uniformly bounded over layers \(1\leq\ell\leq L\).
\end{theorem}

\begin{proposition}[Layerwise readout]
\label{prop:ridge}
For $\RidgeReg>0$ and any finite $\DesignMat{\LayerIdx}$, the ridge objective
$\NormTwo{\DesignMat{\LayerIdx}\ReadoutWeight{\LayerIdx}-\TargetVec}^{2}
+\RidgeReg\NormTwo{\ReadoutWeight{\LayerIdx}}^{2}$ is strictly convex and
Equation~\eqref{eq:ridge} is its unique minimiser.
\end{proposition}

Proposition~\ref{prop:ridge} is the standard ridge argument and is included
because \LAIR fits one such readout at every depth. It is a property of the
estimator rather than of the architecture.

One component remains, which acts after the readouts rather than inside the
recursion. Aggregating across depth is what makes the model an ensemble, and
it is worth being precise about what that does and does not buy.

\begin{proposition}[Aggregation]
\label{prop:aggregation}
Let $\varepsilon^{(1)},\ldots,\varepsilon^{(L)}$ denote the layerwise prediction errors. Define the mean-aggregated error by
\(\bar{\varepsilon} = \frac{1}{L}\sum_{\ell=1}^L\varepsilon^{(\ell)}\). Then
\(\operatorname{Var}(\bar{\varepsilon}) = \frac{1}{L^2}\sum_{i=1}^L\sum_{j=1}^L
\operatorname{Cov}\left(\varepsilon^{(i)},\varepsilon^{(j)}\right)\). If the layerwise errors are uncorrelated in pairs and have a common variance $\sigma^2$, then \(\operatorname{Var}(\bar{\varepsilon}) = \frac{\sigma^2}{L}\). For the median aggregate, consider a fixed realization of the layerwise
errors. If more than half of the layerwise errors satisfy \(|\varepsilon^{(\ell)}|\leq\tau\), then the median-aggregated error satisfies \(
|\varepsilon_{\mathrm{med}}|\leq\tau\).
\end{proposition}

The first part of Proposition~\ref{prop:aggregation} states a condition rather
than a property of \LAIR. The alignment impulse pulls every layer towards the
same anchor, so layerwise errors are correlated by construction and the
variance reduction available in practice is smaller than $\NumLayers^{-1}$.

\section{Experiments}
\label{sec:experiments}

This section reports what \LAIR does on real data. We compare it twice, once
against the randomized-network family it belongs to and once against
conventional models that fit every parameter by iterative optimization. These
are separate questions. The first asks whether the alignment anchor improves on
the randomized recursion it is added to. The second asks whether the resulting
model is competitive outside that family at all. The experiment protocol, dataset details, and extended results are given in Appendix~\ref{app:nonrand}

\subsection{Comparison with randomized networks}

Table~\ref{tab:benchmark_rand} reports the comparison of record. The baselines
are \RVFL, \dRVFL, \edRVFL, \edRVFLSC, Fuzzy \RVFL, and bagged and boosted
\RVFL. \LAIR attains the lowest mean error on $13$ of the $23$ datasets and an average
rank of $1.96$ of eight models. Every baseline is rejected against it after Holm
correction, with the largest adjusted $p$-value $0.0045$.

\begin{table*}[t]
\centering
\caption{Test mean squared error against the randomized-network family, mean $\pm$ standard deviation over ten folds. Lower is better and the best entry in each row is in bold. The final rows give the average rank within this family and the Holm-adjusted Wilcoxon signed-rank $p$-value against \LAIR.}
\label{tab:benchmark_rand}
\resizebox{\textwidth}{!}{%
\begin{tabular}{lcccccccc}
\toprule
Dataset & RVFL & dRVFL & edRVFL & edRVFL-SC & Fuzzy RVFL & RVFL-Bag & RVFL-Boost & LAIR-Net \\
\midrule
Airfoil & 7.943 \tiny{$\pm$ 2.342} & 4.128 \tiny{$\pm$ 1.060} & 5.589 \tiny{$\pm$ 1.397} & 3.391 \tiny{$\pm$ 0.952} & 17.553 \tiny{$\pm$ 3.613} & 10.315 \tiny{$\pm$ 1.574} & 6.716 \tiny{$\pm$ 1.462} & \textbf{2.208 \tiny{$\pm$ 0.485}} \\
Auto MPG & 7.368 \tiny{$\pm$ 1.936} & 7.096 \tiny{$\pm$ 1.885} & 7.613 \tiny{$\pm$ 1.886} & 7.407 \tiny{$\pm$ 2.537} & 8.520 \tiny{$\pm$ 2.138} & 7.160 \tiny{$\pm$ 2.107} & 6.975 \tiny{$\pm$ 1.662} & \textbf{6.919 \tiny{$\pm$ 2.074}} \\
Autos & 0.020 \tiny{$\pm$ 0.006} & 0.021 \tiny{$\pm$ 0.005} & 0.019 \tiny{$\pm$ 0.006} & 0.019 \tiny{$\pm$ 0.006} & 0.025 \tiny{$\pm$ 0.007} & \textbf{0.017 \tiny{$\pm$ 0.007}} & 0.020 \tiny{$\pm$ 0.008} & 0.021 \tiny{$\pm$ 0.010} \\
Breast Cancer & \textbf{869.556 \tiny{$\pm$ 310.974}} & 896.296 \tiny{$\pm$ 291.201} & 917.961 \tiny{$\pm$ 268.373} & 901.050 \tiny{$\pm$ 317.573} & 877.582 \tiny{$\pm$ 279.775} & 896.660 \tiny{$\pm$ 290.495} & 908.803 \tiny{$\pm$ 307.653} & 871.580 \tiny{$\pm$ 265.293} \\
Challenger & 0.495 \tiny{$\pm$ 0.470} & 0.466 \tiny{$\pm$ 0.490} & 0.975 \tiny{$\pm$ 1.160} & 0.510 \tiny{$\pm$ 0.543} & 0.418 \tiny{$\pm$ 0.433} & 0.480 \tiny{$\pm$ 0.415} & \textbf{0.329 \tiny{$\pm$ 0.252}} & 0.389 \tiny{$\pm$ 0.448} \\
Concrete & 56.708 \tiny{$\pm$ 10.841} & 56.480 \tiny{$\pm$ 11.082} & 48.850 \tiny{$\pm$ 9.629} & 50.208 \tiny{$\pm$ 11.808} & 92.994 \tiny{$\pm$ 13.458} & 55.531 \tiny{$\pm$ 11.564} & 54.718 \tiny{$\pm$ 11.492} & \textbf{40.199 \tiny{$\pm$ 10.883}} \\
Concrete Slump & 17.497 \tiny{$\pm$ 11.093} & 26.770 \tiny{$\pm$ 28.230} & 18.699 \tiny{$\pm$ 12.456} & 12.752 \tiny{$\pm$ 9.866} & 53.202 \tiny{$\pm$ 41.126} & 425.757 \tiny{$\pm$ 169.588} & 13.700 \tiny{$\pm$ 13.103} & \textbf{11.170 \tiny{$\pm$ 10.246}} \\
Energy & 0.264 \tiny{$\pm$ 0.101} & 0.224 \tiny{$\pm$ 0.065} & 0.278 \tiny{$\pm$ 0.112} & 0.210 \tiny{$\pm$ 0.068} & 8.105 \tiny{$\pm$ 2.161} & 0.401 \tiny{$\pm$ 0.105} & 0.232 \tiny{$\pm$ 0.062} & \textbf{0.163 \tiny{$\pm$ 0.045}} \\
Fertility & 0.034 \tiny{$\pm$ 0.015} & 0.042 \tiny{$\pm$ 0.024} & \textbf{0.030 \tiny{$\pm$ 0.013}} & 0.036 \tiny{$\pm$ 0.014} & 0.032 \tiny{$\pm$ 0.015} & 0.035 \tiny{$\pm$ 0.017} & 0.036 \tiny{$\pm$ 0.021} & 0.033 \tiny{$\pm$ 0.019} \\
Forest Fires & 2.084 \tiny{$\pm$ 0.599} & 2.074 \tiny{$\pm$ 0.572} & 2.089 \tiny{$\pm$ 0.570} & 2.074 \tiny{$\pm$ 0.566} & 2.095 \tiny{$\pm$ 0.566} & \textbf{1.948 \tiny{$\pm$ 0.471}} & 2.062 \tiny{$\pm$ 0.583} & 2.073 \tiny{$\pm$ 0.582} \\
Gas & 2.93e3 \tiny{$\pm$ 9.28e3} & 20.358 \tiny{$\pm$ 63.440} & 1.86e3 \tiny{$\pm$ 5.90e3} & \textbf{0.314 \tiny{$\pm$ 0.684}} & 181.408 \tiny{$\pm$ 573.123} & 1.36e3 \tiny{$\pm$ 4.29e3} & 52.645 \tiny{$\pm$ 164.803} & 8.752 \tiny{$\pm$ 27.367} \\
Housing & 12.606 \tiny{$\pm$ 9.308} & 11.514 \tiny{$\pm$ 8.647} & 11.983 \tiny{$\pm$ 11.302} & \textbf{10.109 \tiny{$\pm$ 9.456}} & 21.788 \tiny{$\pm$ 9.876} & 10.158 \tiny{$\pm$ 8.853} & 11.468 \tiny{$\pm$ 9.432} & 10.706 \tiny{$\pm$ 8.377} \\
Machine CPU & 0.161 \tiny{$\pm$ 0.037} & 0.180 \tiny{$\pm$ 0.088} & 0.175 \tiny{$\pm$ 0.051} & 0.162 \tiny{$\pm$ 0.048} & 0.191 \tiny{$\pm$ 0.057} & 0.165 \tiny{$\pm$ 0.043} & 0.152 \tiny{$\pm$ 0.039} & \textbf{0.146 \tiny{$\pm$ 0.034}} \\
Parkinsons & 83.880 \tiny{$\pm$ 5.141} & 58.557 \tiny{$\pm$ 17.420} & 78.320 \tiny{$\pm$ 7.886} & 55.378 \tiny{$\pm$ 20.058} & 83.235 \tiny{$\pm$ 3.515} & 54.792 \tiny{$\pm$ 3.869} & 93.392 \tiny{$\pm$ 8.443} & \textbf{24.092 \tiny{$\pm$ 17.750}} \\
Pendulum & 3.282 \tiny{$\pm$ 1.610} & 2.780 \tiny{$\pm$ 1.230} & 2.595 \tiny{$\pm$ 1.037} & 2.467 \tiny{$\pm$ 1.060} & 7.393 \tiny{$\pm$ 3.498} & 2.269 \tiny{$\pm$ 1.183} & 2.798 \tiny{$\pm$ 1.809} & \textbf{1.681 \tiny{$\pm$ 0.963}} \\
PumaDyn-32nm & 1.001 \tiny{$\pm$ 0.066} & 0.987 \tiny{$\pm$ 0.062} & 0.993 \tiny{$\pm$ 0.067} & 0.954 \tiny{$\pm$ 0.060} & 1.004 \tiny{$\pm$ 0.066} & 0.947 \tiny{$\pm$ 0.062} & 0.987 \tiny{$\pm$ 0.064} & \textbf{0.052 \tiny{$\pm$ 0.004}} \\
Servo & 0.095 \tiny{$\pm$ 0.056} & 0.102 \tiny{$\pm$ 0.051} & 0.092 \tiny{$\pm$ 0.044} & 0.103 \tiny{$\pm$ 0.051} & 0.219 \tiny{$\pm$ 0.088} & 0.153 \tiny{$\pm$ 0.060} & 0.095 \tiny{$\pm$ 0.051} & \textbf{0.086 \tiny{$\pm$ 0.047}} \\
SkillCraft & 0.070 \tiny{$\pm$ 0.019} & 0.064 \tiny{$\pm$ 0.008} & 0.067 \tiny{$\pm$ 0.013} & 0.075 \tiny{$\pm$ 0.034} & 0.072 \tiny{$\pm$ 0.026} & 0.063 \tiny{$\pm$ 0.009} & \textbf{0.063 \tiny{$\pm$ 0.009}} & 0.066 \tiny{$\pm$ 0.011} \\
SML & 1.219 \tiny{$\pm$ 0.106} & 0.246 \tiny{$\pm$ 0.052} & 0.907 \tiny{$\pm$ 0.050} & 0.356 \tiny{$\pm$ 0.022} & 2.044 \tiny{$\pm$ 0.138} & 0.885 \tiny{$\pm$ 0.101} & 0.736 \tiny{$\pm$ 0.040} & \textbf{0.117 \tiny{$\pm$ 0.023}} \\
Solar & 0.627 \tiny{$\pm$ 0.344} & 0.625 \tiny{$\pm$ 0.342} & 0.614 \tiny{$\pm$ 0.335} & 0.619 \tiny{$\pm$ 0.344} & 0.624 \tiny{$\pm$ 0.358} & \textbf{0.612 \tiny{$\pm$ 0.345}} & 0.621 \tiny{$\pm$ 0.348} & 0.622 \tiny{$\pm$ 0.348} \\
Stock & 2.63e-5 \tiny{$\pm$ 4.84e-6} & 2.68e-5 \tiny{$\pm$ 5.05e-6} & 2.59e-5 \tiny{$\pm$ 5.34e-6} & 2.59e-5 \tiny{$\pm$ 5.76e-6} & \textbf{2.53e-5 \tiny{$\pm$ 4.91e-6}} & 2.84e-5 \tiny{$\pm$ 4.92e-6} & 2.58e-5 \tiny{$\pm$ 4.90e-6} & 2.56e-5 \tiny{$\pm$ 3.88e-6} \\
Wine Quality & 0.272 \tiny{$\pm$ 0.031} & 0.251 \tiny{$\pm$ 0.037} & 0.240 \tiny{$\pm$ 0.032} & 0.225 \tiny{$\pm$ 0.036} & 0.330 \tiny{$\pm$ 0.060} & 0.251 \tiny{$\pm$ 0.040} & 0.236 \tiny{$\pm$ 0.032} & \textbf{0.218 \tiny{$\pm$ 0.038}} \\
Yacht & 0.036 \tiny{$\pm$ 0.069} & 0.024 \tiny{$\pm$ 0.036} & 0.029 \tiny{$\pm$ 0.049} & 0.024 \tiny{$\pm$ 0.040} & 0.141 \tiny{$\pm$ 0.142} & 0.073 \tiny{$\pm$ 0.064} & 0.032 \tiny{$\pm$ 0.043} & \textbf{0.020 \tiny{$\pm$ 0.032}} \\
\midrule
Average rank & 5.65 & 4.78 & 4.87 & 3.70 & 6.61 & 4.43 & 4.00 & 1.96 \\
Holm-adjusted $p$ & 0.0014 & 0.0003 & 0.0007 & 0.0045 & 0.0003 & 0.0045 & 0.0029 & -- \\
\bottomrule
\end{tabular}}
\end{table*}
% END INLINE: tables/benchmark_randomized.tex

\begin{figure}[t]
\centering
\includegraphics[width=0.88\columnwidth]{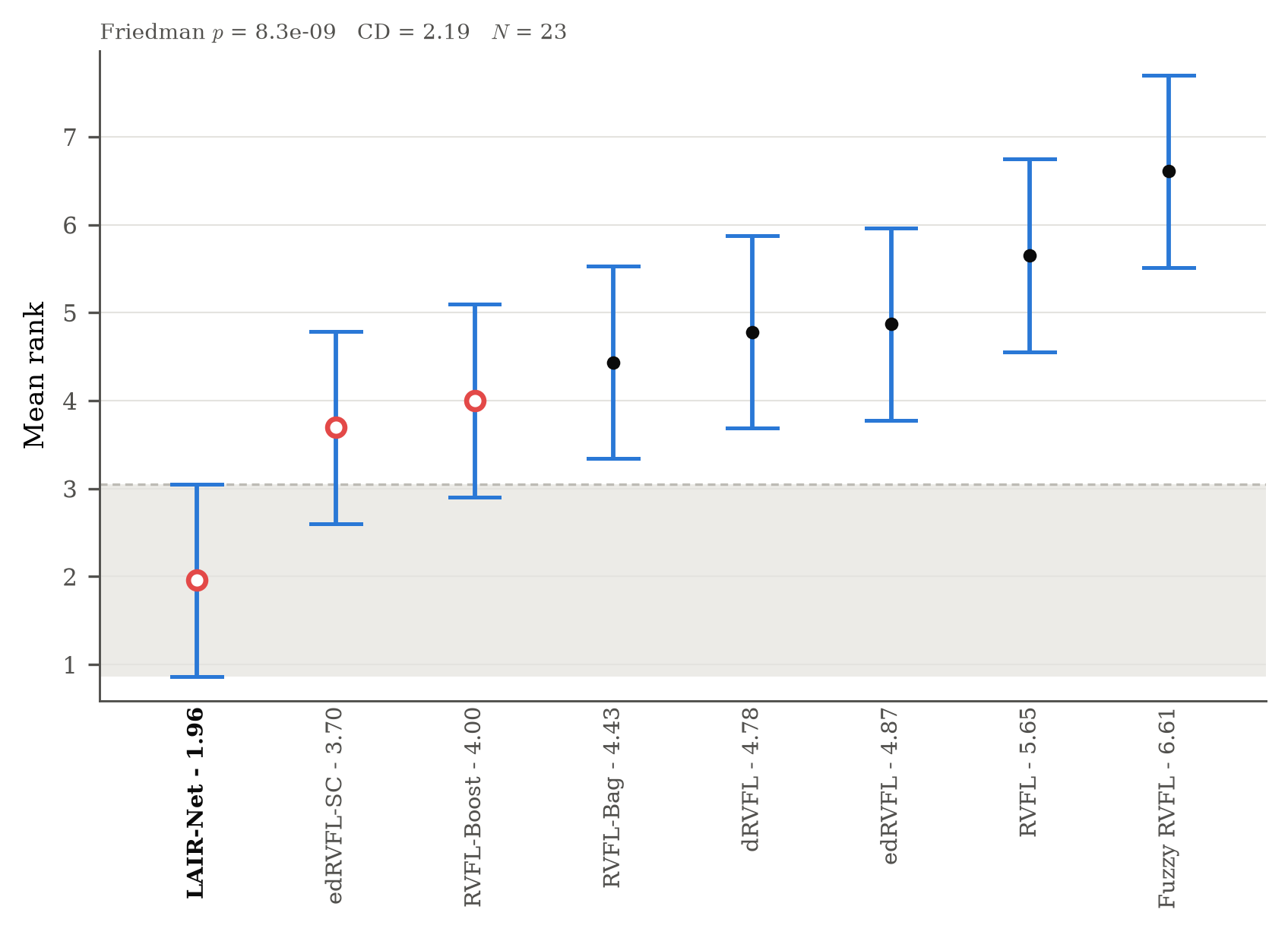}
\caption{Nemenyi comparison within the randomized-network family. Each point is
a mean rank with its interval, and the shaded band marks the region not
separated from the best-ranked model at the $0.05$ level, with
$\mathrm{CD}=2.19$ over $23$ datasets. Lower is better.}
\label{fig:nemenyi_rand}
\end{figure}

Figure~\ref{fig:nemenyi_rand} shows where that ranking is and is not decisive.
The critical difference separates \LAIR from five of the seven baselines, namely
bagged \RVFL, \dRVFL, \edRVFL, \RVFL, and Fuzzy \RVFL. It does not separate \LAIR
from \edRVFLSC or from boosted \RVFL, whose intervals overlap the band.

% END INLINE: draft_experiments.tex

% Summary of the controlled study; complete design and results are in the appendix.
\section{Simulation study}
\label{sec:simulation}

We vary the nonlinear share of a regression target and its noise level
independently to test when the alignment anchor improves on a randomized
baseline. With $1000$ training and $1000$ test samples per seed, the
\edRVFL-to-\LAIR test-error ratio rises as nonlinear signal strengthens and
falls as noise increases. At the fixed alignment coefficient
$\AlignCoef=0.5$, \LAIR overtakes \edRVFL in the low-noise setting once
nonlinear structure is present, while it does not overtake \edRVFL within
the tested nonlinear-share grid at the highest noise level. The complete data-generating process, grid, and confidence intervals are reported in Appendix~\ref{app:simulation}.

Removing only the supervised anchor leaves a residual and leaky recursion whose error differs from \edRVFL by at most $3.7\%$ across the fifteen tested cells. The anchor accounts for almost all of the observed gain; capacity and untrained-anchor controls do not close that gap in the
high-signal regime. Selecting $\AlignCoef$ on a validation split gives
test error within $1.8\%$ of the best value on the grid in every cell and
reduces the cells where \LAIR loses to \edRVFL from nine of fifteen
with the fixed coefficient to one of fifteen. At $250$ training samples,
the direction persists, but only the low-noise gains remain clearly
separated. Appendix~\ref{app:ablation} gives the ablations and
Appendix~\ref{app:simulation} sensitivity analyses.

Across the $23$ benchmark datasets, a proxy for the reward from nonlinearity has a rank correlation of $-0.598$ ($p=0.003$) with \LAIR's log error ratio against the best randomized baseline; a noise proxy has a correlation of $+0.425$ ($p=0.043$). These are observational associations across datasets, consistent with the controlled study rather than causal evidence. The proxy definitions, plot, and outlier analysis appear in Appendix~\ref{app:simulation}.

\section{Conclusion}
\label{sec:conclusion}
\label{sec:limitations}

\LAIR combines a learned target-aware alignment anchor with leaky residual transitions while retaining closed-form readouts and fixed hidden width across depth. It achieves the best average ranks among eight randomized models ($1.96$) and twelve conventional models ($3.26$), with Holm-adjusted tests favoring it over every baseline; Nemenyi tests separate it from five randomized and six conventional baselines. Controlled experiments attribute most of the improvement to the alignment anchor, whose benefit depends on the learnable nonlinear structure and diminishes for nearly linear targets or dominant noise, consistent with associations observed across benchmark datasets. Validation-based selection of the alignment coefficient reduces underperformance against the randomized baseline from nine of fifteen simulation settings to one. Limitations include smaller gains with limited samples, unresolved behavior in high-dimensional mixed-feature settings, vulnerability to out-of-range inputs, and several-fold higher training cost than \edRVFL due to anchor fitting. Theorem~\ref{thm:input-perturbation} applies only under $\PropFac<1$ and characterizes propagation across depth rather than explaining empirical performance. Although the direction of the nonlinear-structure effect is stable across tested capacities, its crossover threshold is not. Future work should estimate the anchor's advantage over the randomized ensemble directly to guide alignment strength and remove the corresponding search hyperparameter.

\begingroup
\sloppy
\emergencystretch=3em
\printbibliography
\endgroup
\clearpage
\clearpage
\appendix
\thispagestyle{empty}

% Supplementary material: To improve readability, you must use a single-column format for the supplementary material.
\onecolumn
\section*{Appendix}
\section{Simulation study}
\label{app:simulation}

The benchmark comparison in Section~\ref{sec:experiments} reports where \LAIR
performs well, but it does not identify the data-generating reason, because a
real dataset does not come with a knob for the property the architecture is
meant to exploit. This section supplies that knob. We construct a
data-generating process in which the amount
of nonlinear structure and the noise level are set independently, and we use it
to answer three questions. Under which conditions does a difference from a
randomized baseline appear, which component of the architecture produces it,
and does the resulting account carry over to real data.

Covariates are drawn in three blocks, $\NumFeatures=30$ in total, comprising ten
informative Gaussian variables $\DgpLatent$, ten redundant variables formed as
noisy linear images of $\DgpLatent$, and ten pure-noise variables. The response
combines a linear and a nonlinear component,

\begin{equation}
  y=\sqrt{1-\NlShare}\;\LinSig(\DgpLatent)
   +\sqrt{\NlShare}\;\NlSig(\DgpLatent)+\NoiseSd\,\epsilon ,
  \qquad \epsilon\sim\mathcal{N}(0,1),
  \label{eq:dgp}
\end{equation}

where $\LinSig$ is linear in $\DgpLatent$ and $\NlSig$ combines quadratic,
interaction and trigonometric terms. Both components are standardised to unit
variance, so $\NlShare$ is the nonlinear share of the signal variance and
$\NoiseSd$ fixes the noise level independently of $\NlShare$. The realised
nonlinear share tracks the requested value closely, taking the values $0.000$,
$0.261$, $0.526$, $0.783$ and $1.000$ for
$\NlShare\in\{0,0.25,0.5,0.75,1\}$.

Equation~\eqref{eq:dgp} isolates the property at issue. Increasing $\NlShare$
adds structure that a linear readout on the raw inputs does not represent
exactly, while holding the number of variables, their correlation structure and
the noise level fixed. Every comparison below uses $\NumSamples=1000$ training and $1000$ test
samples, ten seeds, and matched hidden width, depth, activation, input scale and
ridge penalty across all models. Hyperparameters are fixed rather than tuned, so
that a difference between two arms can only come from the component that differs
between them.

Baselines are drawn from the randomized family alone, namely \RVFL, \dRVFL,
\edRVFL and \edRVFLSC. A cell is reported as a win or a loss only when the
$95\%$ bootstrap confidence interval of the median mean-squared-error ratio
across seeds excludes one, and crossovers are therefore reported as regions
rather than thresholds.

\subsection{Where the advantage appears}

\begin{figure}[htbp]
\centering
\includegraphics[width=0.86\columnwidth]{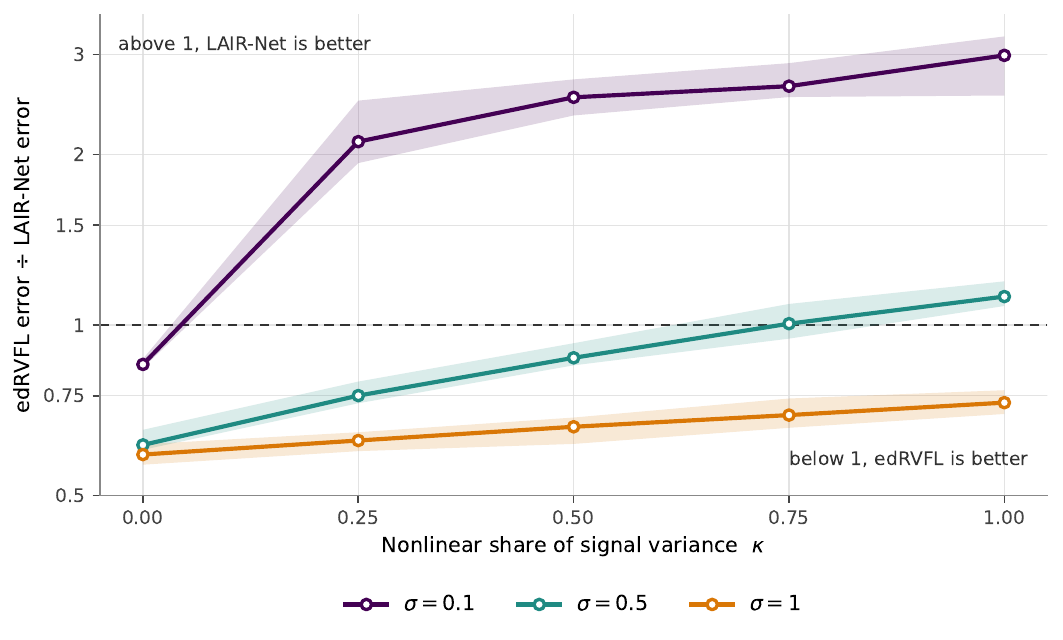}
\caption{Where the advantage appears. Each curve is the median ratio of \edRVFL
to \LAIR test error over ten seeds at one noise level, with the alignment
coefficient fixed at $\AlignCoef=0.5$, and the band is a $95\%$ percentile
bootstrap over those seeds. The dashed line at one is the crossover. At
$\NoiseSd=0.1$ the curve crosses between $\NlShare=0$ and $\NlShare=0.25$, at
$\NoiseSd=0.5$ it reaches one only at $\NlShare=0.75$, and at $\NoiseSd=1.0$ it
does not cross within the grid.}
\label{fig:crossover}
\end{figure}

Figure~\ref{fig:crossover} and Table~\ref{tab:crossover} report the ratio of
\edRVFL to \LAIR test error with the alignment coefficient held at
$\AlignCoef=0.5$. The ratio increases monotonically with $\NlShare$ at every
noise level and decreases with $\NoiseSd$ at every nonlinear share, and the
figure shows where each curve meets one, which is the quantity the rest of this
section is about.

\begin{table}[htbp]
\centering
\caption{Median ratio of \edRVFL to \LAIR test mean squared error over ten
seeds, with $\AlignCoef=0.5$ fixed. A ratio above one favours \LAIR. Rows vary
the nonlinear share $\NlShare$ of Equation~\eqref{eq:dgp}; columns vary the
noise level $\NoiseSd$.}
\label{tab:crossover}
\begin{tabular}{lccc}
\toprule
$\NlShare$ & $\NoiseSd=0.1$ & $\NoiseSd=0.5$ & $\NoiseSd=1.0$ \\
\midrule
0.00 & 0.852 & 0.614 & 0.591 \\
0.25 & 2.106 & 0.751 & 0.625 \\
0.50 & 2.522 & 0.875 & 0.661 \\
0.75 & 2.639 & 1.005 & 0.693 \\
1.00 & 2.991 & 1.122 & 0.729 \\
\bottomrule
\end{tabular}
\end{table}

The crossover from loss to gain occurs for
$\NlShare\in(0,0.25]$ at $\NoiseSd=0.1$ and for
$\NlShare\in(0.5,1.0]$ at $\NoiseSd=0.5$, and does not occur within
$\NlShare\leq 1$ at $\NoiseSd=1.0$. \LAIR therefore helps when the target has
nonlinear structure that is learnable at the available noise level, and the
randomized baselines are preferable when the target is close to linear or when
noise dominates. An independent reimplementation of the design agreed to within
three per cent on every entry of
Table~\ref{tab:crossover}.

\subsection{Which component produces the difference}

The comparison in Table~\ref{tab:crossover} varies three things at once, since
\LAIR differs from \edRVFL in the residual layer input, the leaky transition and
the alignment impulse. Inserting an arm with $\AlignCoef=0$, which retains the
recursion and removes only the anchor, separates them.

\begin{figure}[t]
\centering
\includegraphics[width=\columnwidth]{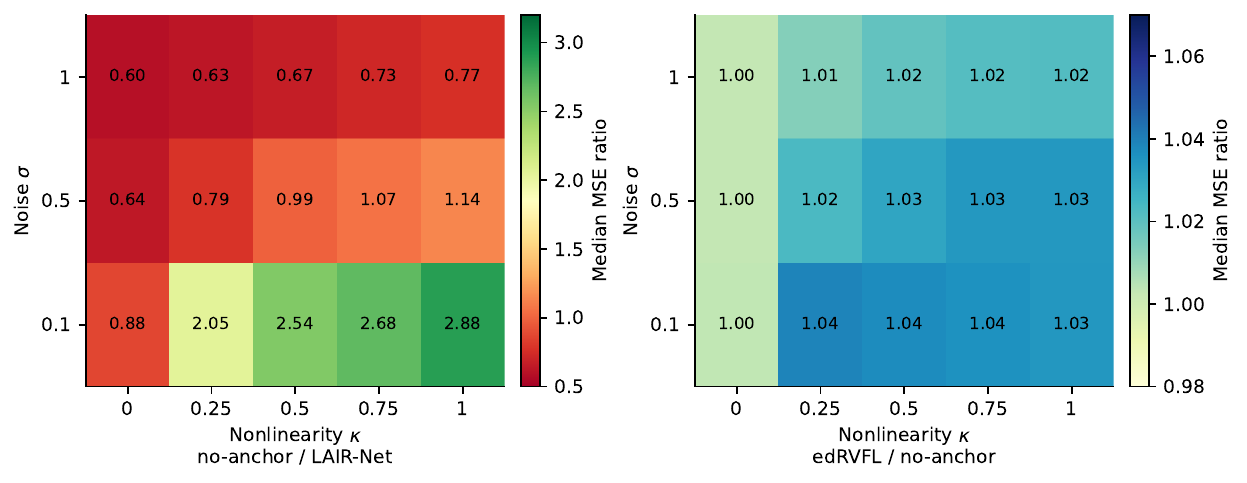}
\caption{Decomposition of the \edRVFL-to-\LAIR error ratio across the
$(\NlShare,\NoiseSd)$ grid. The architecture term compares \edRVFL with an
$\AlignCoef=0$ arm that keeps the residual layer input and the leaky transition;
the anchor term compares that arm with the full model. Values above one favour
the model on the right of each comparison.}
\label{fig:decomposition}
\end{figure}

The architecture term lies between $1.002$ and $1.037$ across all fifteen cells,
so the recursion without an anchor differs from \edRVFL by at most $3.7\%$. The
anchor term reproduces the pattern of Table~\ref{tab:crossover}, reaching
$2.889$ at $\NlShare=1.0$, $\NoiseSd=0.1$. The alignment anchor accounts for
essentially all of the difference, and the randomized recursion on its own
accounts for little of it.

We may reasonably object that the anchor is a trained network, so the
gain could reflect additional trained capacity rather than alignment. Two
controls address this. Giving the randomized baselines up to eight times \LAIR's
hidden width leaves the best of them at $3.111$ times \LAIR's error in the
high-signal cell, and repeating that sweep over six ridge penalties with the
penalty selected on the test set, which is an advantage the baseline would not
have in practice, leaves it at $3.084$. Extra capacity does not close the gap in
this regime. A third control replaces the trained anchor with an untrained
network of identical width and distribution, which removes the target
information while retaining the shrinkage the mixing step applies. That arm buys
nothing at any sample size tested, so the contribution is target-aware
information rather than generic shrinkage.

The results above hold $\AlignCoef$ fixed, and the value that is best depends on
$\NlShare$ and $\NoiseSd$, neither of which is observable on a real dataset. The
practical question is therefore whether the model's own validation signal
selects a suitable $\AlignCoef$ without being told the regime.

\begin{figure}[t]
\centering
\includegraphics[width=\columnwidth]{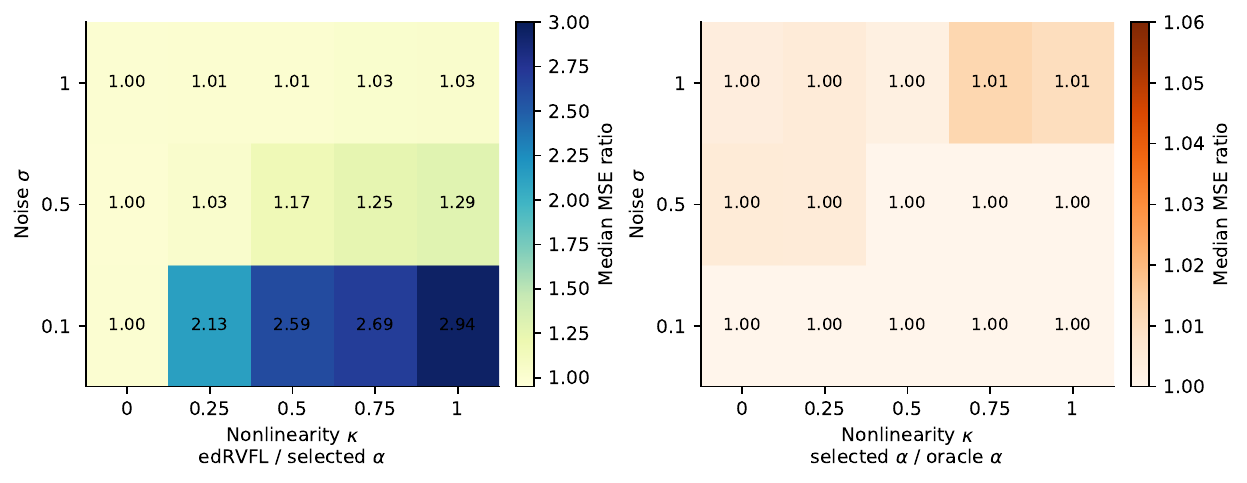}
\caption{Effect of selecting $\AlignCoef$ on a held-out validation split, with
all other hyperparameters fixed. Left, the \edRVFL-to-\LAIR error ratio with
$\AlignCoef$ selected; right, the ratio of the selected model's error to that of
the best $\AlignCoef$ in the grid. A ratio near one on the right indicates that
validation recovered the best available setting.}
\label{fig:alpha-selection}
\end{figure}

\begin{figure}[t]
\centering
\includegraphics[width=\columnwidth]{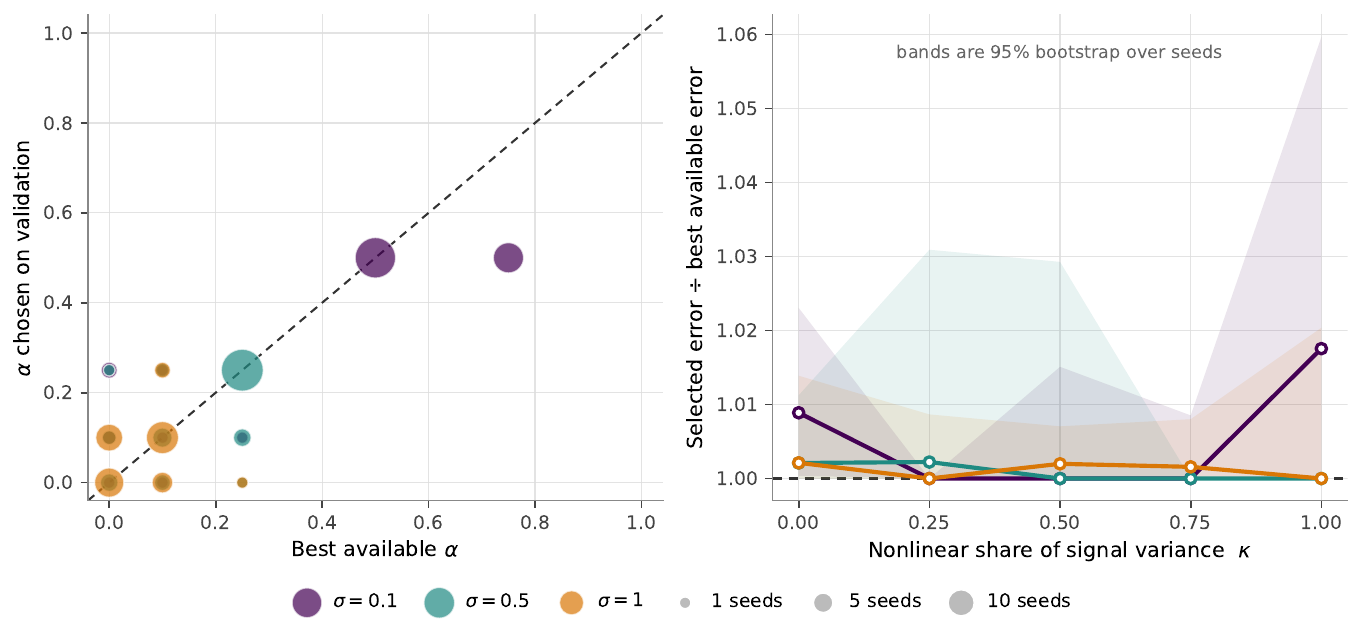}
\caption{What validation selection of $\AlignCoef$ recovers, and what it costs.
Left, the selected $\AlignCoef$ against the best available one, with marker area
proportional to the number of seeds landing on that pair and the dashed line
marking exact agreement. Right, the ratio of the selected model's error to the
error at the best available $\AlignCoef$, by nonlinear share, with $95\%$
percentile bootstrap bands over seeds. Points on the dashed line on the right
lose nothing by selecting.}
\label{fig:alpha-recovery}
\end{figure}

Selecting $\AlignCoef$ from a six-point grid on a validation split recovers the
best available setting almost exactly. The ratio of the selected model's error
to the best in the grid lies between $1.000$ and $1.018$ across all cells, and
the rank correlation between the selected and the best $\AlignCoef$ is $0.838$.
Figure~\ref{fig:alpha-recovery} shows both quantities directly.

The selected value falls from $0.5$ where the anchor is informative to $0$ on a
purely linear target. Under the interval criterion, fixing $\AlignCoef=0.5$
loses in nine of fifteen cells with a worst case of $0.591$, whereas selecting it
loses in one cell, by $1.1\%$, on the purely linear high-signal problem. The
mechanism degrades gracefully. Where the anchor helps it is used, and where it
does not it is switched off.

\begin{figure}[t]
\centering
\includegraphics[width=\columnwidth]{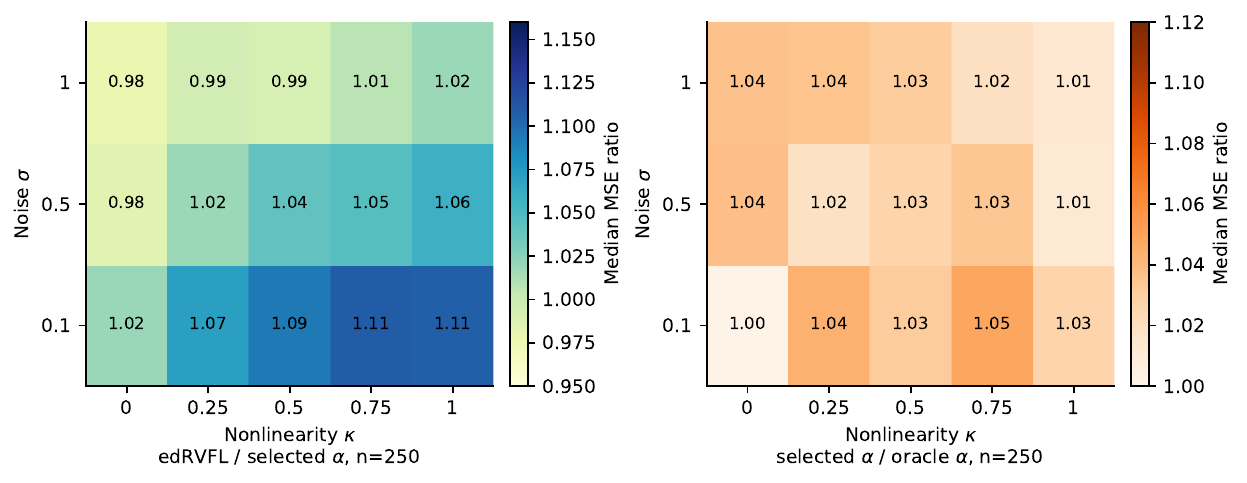}
\caption{Alignment selection at $\NumSamples=250$. Left, the error of the
selected $\AlignCoef$ relative to the best in the grid; right, the
\edRVFL-to-\LAIR ratio at the best $\AlignCoef$. Effect sizes are smaller than
at $\NumSamples=1000$ and the low-noise cells retain the clearest separation.}
\label{fig:small-n}
\end{figure}

At $\NumSamples=250$ the same qualitative pattern holds, with the selected model
remaining close to the best available setting and the best setting still ahead of
\edRVFL in every cell, but effect sizes shrink and the low-noise advantage is
the only part that remains clearly separated. Figure~\ref{fig:small-n} reports
the detail.

The grid above uses one setting of hidden width, input scale and ridge penalty.
Repeating it at a narrower setting and a wider one leaves the ordering
unchanged. The error ratio is non-decreasing in $\NlShare$ at every noise level
for all three settings, and the high-signal crossover remains within
$\NlShare\in(0,0.25]$. The crossover at $\NoiseSd=0.5$ moves, appearing within
$\NlShare\in(0.25,0.5]$ at the wider setting and not appearing at all at the
narrower one. The qualitative account is therefore stable across these settings
while the exact crossover location is not, and we report the former as the
finding and the latter as a caveat.

\subsection{Does the account transfer to real data}
\label{subsec:transfer}

The account so far is built on simulated data, where both axes are set by us.
The test that matters is whether it describes the benchmark results of
Section~\ref{sec:experiments}, where neither axis is observable. Both have
computable proxies. We estimate the reward for nonlinearity as
$\NlShareHat=\log(\MSE_{\text{linear}}/\MSE_{\text{best}})$ and the noise floor
as $\NoiseSdHat=\MSE_{\text{best}}/\VarOp(y)$, using only error values already
computed for that section, with no additional fitting. If the mechanism
transfers, \LAIR should do relatively better on datasets with large
$\NlShareHat$ and small $\NoiseSdHat$. The outcome variable is the ratio of
\LAIR's error to that of the best randomized baseline on the same dataset.

\begin{figure}[t]
\centering
\includegraphics[width=0.85\columnwidth]{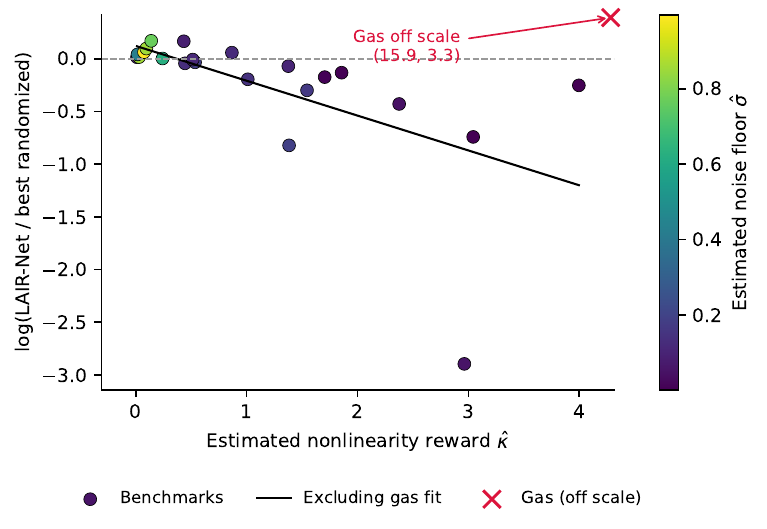}
\caption{Relative performance of \LAIR against the best randomized baseline on
each benchmark dataset, plotted against the nonlinearity proxy $\NlShareHat$.
The vertical axis is $\log(\MSE_{\LAIR}/\MSE_{\text{best randomized}})$, so
points below the horizontal line at zero favour \LAIR. The gas dataset is marked
as an off-scale point and is discussed in the text.}
\label{fig:transfer}
\end{figure}

The association predicted by the simulation is present. The rank correlation
between $\NlShareHat$ and the log error ratio is $-0.598$ ($p=0.003$), so
datasets that reward nonlinearity are associated with better relative
performance, and the correlation with $\NoiseSdHat$ is $+0.425$ ($p=0.043$), so
noisier datasets are associated with worse relative performance. Splitting at
the median of $\NlShareHat$, \LAIR improves on the best randomized baseline on
ten of eleven datasets above the median and on three of twelve below it.

These are observational associations across datasets, not causal statements. The
proxies are computed from the same error values that enter the outcome, and many
dataset properties covary with nonlinearity, so the result should be read as
evidence that the simulation's account is consistent with the benchmark
behaviour rather than as a demonstration that nonlinear structure causes the
improvement.

One dataset does not follow the pattern. On gas, $\NlShareHat$ is large but
\LAIR performs poorly in both absolute and relative terms. Inspecting the folds
shows a test sample whose value on one feature lies far outside the training
range for that feature, which affects every model fitted on standardised inputs
and affects those with unbounded extrapolation most. We identify it as an
out-of-support case rather than an instance of the mechanism above, and we note
that excluding it strengthens both correlations, to $-0.826$ and $+0.554$
respectively. We report the correlations with gas included as the headline
values and the excluding-gas values as a sensitivity check, rather than the
reverse.

\section{Proofs}
\label{app:proofs}
\suppressfloats[t]

Throughout, $\LayerInputVec{\LayerIdx}=[1,\HiddenVec{\LayerIdx-1},\InputVec]$
and $\RandWeight{\LayerIdx}$ is partitioned row-wise into the bias row, the
state block $\RandWeightHj{\LayerIdx}$ and the input block
$\RandWeightXj{\LayerIdx}$, so that
\[
  (\RandWeight{\LayerIdx})\Transpose\LayerInputVec{\LayerIdx}
  =(\RandWeightBXj{\LayerIdx})\Transpose[1,\InputVec]
  +(\RandWeightHj{\LayerIdx})\Transpose\HiddenVec{\LayerIdx-1}.
\]
The constants $\StateGain$, $\InputGain$ and $\BiasInputGain$ are the
depth-uniform quantities of Equations~\eqref{eq:constants}
and~\eqref{eq:biasgain}, so a bound involving them holds at every layer with the
same value and may be iterated across depth.

\subsection*{Proof of Theorem~\ref{thm:input-perturbation}}

Let the two trajectories be generated from $x$ and $\bar{x}$ with
anchors $s$ and $\bar{s}$ under the same realization of the random
weights. Since the bias component is identical for both trajectories,
the layer inputs satisfy
\[
z^{(\ell)}-\bar{z}^{(\ell)} = [0,\Delta h^{(\ell-1)},\Delta x].
\]
Hence,
\[
(W^{(\ell)})^\top \left(z^{(\ell)}-\bar{z}^{(\ell)}\right) = (W_h^{(\ell)})^\top\Delta h^{(\ell-1)} + (W_x^{(\ell)})^\top\Delta x.
\]
By the Lipschitz property of $g$, the triangle inequality, and the
definitions of $c_1$ and $c_x$,
\[
\begin{aligned}
\|\Delta\widetilde{h}^{(\ell)}\|_2
&\leq
L_g
\left\|
(W_h^{(\ell)})^\top\Delta h^{(\ell-1)} + (W_x^{(\ell)})^\top\Delta x
\right\|_2\\
&\leq
c_1\|\Delta h^{(\ell-1)}\|_2 + c_x\|\Delta x\|_2.
\end{aligned}
\]
The leaky update gives \(\Delta u^{(\ell)} = \gamma\Delta\widetilde{h}^{(\ell)} + (1-\gamma)\Delta h^{(\ell-1)}\), and therefore
\[
\|\Delta u^{(\ell)}\|_2
\leq
(\gamma c_1+1-\gamma)
\|\Delta h^{(\ell-1)}\|_2
+
\gamma c_x\|\Delta x\|_2.
\]
The alignment step satisfies \(\Delta h^{(\ell)} = (1-\alpha)\Delta u^{(\ell)} + \alpha\Delta s\). Consequently,
\[
\|\Delta h^{(\ell)}\|_2
\leq
(1-\alpha)(\gamma c_1+1-\gamma)
\|\Delta h^{(\ell-1)}\|_2
+
(1-\alpha)\gamma c_x\|\Delta x\|_2
+
\alpha\|\Delta s\|_2.
\]
Thus, \(\|\Delta h^{(\ell)}\|_2 \leq \rho\|\Delta h^{(\ell-1)}\|_2+\delta\), where \(\rho=(1-\alpha)(\gamma c_1+1-\gamma)\) and \(\delta = (1-\alpha)\gamma c_x\|\Delta x\|_2 + \alpha\|\Delta s\|_2\).\\
Since $\Delta h^{(0)}=0$, iterating the recursion yields \(
\|\Delta h^{(\ell)}\|_2 \leq \delta\sum_{k=0}^{\ell-1}\rho^k.\) Therefore,
\[
\|\Delta h^{(\ell)}\|_2
\leq
\begin{cases}
\displaystyle
\delta\frac{1-\rho^\ell}{1-\rho},
& \rho\neq1,\\[8pt]
\ell\delta,
& \rho=1.
\end{cases}
\]
When $\rho<1$, \(\|\Delta h^{(\ell)}\|_2 \leq \frac{\delta}{1-\rho}\), which is uniform over $1\leq\ell\leq L$.\\
Finally, if the alignment mapping $x\mapsto s(x)$ is $L_S$-Lipschitz, then \(\|\Delta s\|_2 \leq L_S\|\Delta x\|_2\), and hence
\[
\delta \leq \left[(1-\alpha)\gamma c_x+\alpha L_S
\right]\|\Delta x\|_2.
\]
This completes the proof.
\qed

\subsection*{Proof of Corollary~\ref{cor:state-perturbation}}

Since the two trajectories share the same input and the same anchor, \(\PerturbInput=\mathbf{0}\), \(\PerturbAnchor=\mathbf{0}\), and hence $\delta=0$. Therefore, the recursion of the one-step difference derived in the
proof of Theorem~\ref{thm:input-perturbation} reduces to
\[
\NormTwo{\PerturbHidden{\LayerIdx}}
\leq
\PropFac
\NormTwo{\PerturbHidden{\LayerIdx-1}}.
\]
Iterating this inequality from the initial perturbation gives
\[
\NormTwo{\PerturbHidden{\LayerIdx}}
\leq
\PropFac^{\LayerIdx}
\NormTwo{\PerturbHidden{0}},
\qquad
1\leq\LayerIdx\leq L.
\]
When $\rho<1$, the perturbation therefore decays geometrically with
depth.
\qed

\subsection*{Proof of Theorem~\ref{thm:bounded}}

For a fixed realization of the random weights, write \(z^{(\ell)}=[1,h^{(\ell-1)},x]\). Since $g(0)=0$ and $g$ is $L_g$-Lipschitz, \(\|g(a)\|_2 = \|g(a)-g(0)\|_2 \leq L_g\|a\|_2\). Hence
\[
\|\widetilde h^{(\ell)}\|_2 \leq L_g\left\|(W^{(\ell)})^\top z^{(\ell)}
\right\|_2 \leq L_g \left(\|(W_{0x}^{(\ell)})^\top[1,x]\|_2 + \|(W_h^{(\ell)})^\top h^{(\ell-1)}\|_2\right).
\]
By Assumption 4.1, \(\|[1,x]\|_2 = \sqrt{1+\|x\|_2^2} \leq \sqrt{1+B_x^2}\). Therefore, by the definitions of $c_0$ and $c_1$,
\[
\|\widetilde h^{(\ell)}\|_2
\leq
c_0+c_1\|h^{(\ell-1)}\|_2.
\]
Using the leaky update, \(u^{(\ell)} = \gamma\widetilde h^{(\ell)} + (1-\gamma)h^{(\ell-1)}\), we obtain \(\|u^{(\ell)}\|_2 \leq \gamma c_0 + (\gamma c_1+1-\gamma)\|h^{(\ell-1)}\|_2\). The alignment step gives \(h^{(\ell)} =(1-\alpha)u^{(\ell)} + \alpha s\). Since $\|s\|_2\leq B_s$,
\[
\begin{aligned}
\|h^{(\ell)}\|_2
&\leq (1-\alpha)\|u^{(\ell)}\|_2 + \alpha\|s\|_2\\
&\leq
(1-\alpha)\gamma c_0 + (1-\alpha)(\gamma c_1+1-\gamma)\|h^{(\ell-1)}\|_2 +\alpha B_s.
\end{aligned}
\]
Thus
\(\|h^{(\ell)}\|_2 \leq \rho\|h^{(\ell-1)}\|_2+b\), where \(\rho=(1-\alpha)(\gamma c_1+1-\gamma)\) and \(b=(1-\alpha)\gamma c_0+\alpha B_s\).\\
Since $h^{(0)}=0$, iterating the recursion yields
\[
\|h^{(\ell)}\|_2 \leq b\sum_{k=0}^{\ell-1}\rho^k.
\]
Therefore,
\[
\|h^{(\ell)}\|_2
\leq
\begin{cases}
\displaystyle
b\frac{1-\rho^\ell}{1-\rho},
& \rho\neq1,\\[8pt]
\ell b,
& \rho=1.
\end{cases}
\]
If $\rho<1$, then
\[
\|h^{(\ell)}\|_2 \leq \frac{b}{1-\rho}, \qquad 1\leq\ell\leq L.
\]
Thus the hidden state is uniformly bounded over the layers.

\subsection*{Proof of Proposition~\ref{prop:ridge}}

Let $\ReadoutDim=\NumFeatures+1+\HiddenWidth$ for the readout dimension
and consider
\[
  J(\ReadoutWeight{\LayerIdx})
  =\NormTwo{\DesignMat{\LayerIdx}\ReadoutWeight{\LayerIdx}-\TargetVec}^{2}
  +\RidgeReg\NormTwo{\ReadoutWeight{\LayerIdx}}^{2}.
\]
Its Hessian is $2\bigl[(\DesignMat{\LayerIdx})\Transpose\DesignMat{\LayerIdx}
+\RidgeReg\IdentityMat{\ReadoutDim}\bigr]$. The first term is positive semidefinite and
$\RidgeReg>0$ by Assumption~\ref{as:convex}, so the Hessian is positive definite
and $J$ is strictly convex with a unique stationary point. Setting the gradient
to zero gives the normal equations
\[
  \bigl[(\DesignMat{\LayerIdx})\Transpose\DesignMat{\LayerIdx}
  +\RidgeReg\IdentityMat{\ReadoutDim}\bigr]\ReadoutWeight{\LayerIdx}
  =(\DesignMat{\LayerIdx})\Transpose\TargetVec ,
\]
whose coefficient matrix is invertible, yielding
Equation~\eqref{eq:ridge}. \qed

\subsection*{Proof of Proposition~\ref{prop:aggregation}}

For the mean aggregate, \(\bar{\varepsilon} = \frac{1}{L}\sum_{\ell=1}^L\varepsilon^{(\ell)}\). Therefore, by bilinearity of covariance, 
\[
\begin{aligned}
\operatorname{Var}(\bar{\varepsilon})
&=
\operatorname{Var}
\left(
\frac{1}{L}\sum_{\ell=1}^L\varepsilon^{(\ell)}
\right)\\
&=
\frac{1}{L^2}
\sum_{i=1}^L\sum_{j=1}^L
\operatorname{Cov}
\left(\varepsilon^{(i)},\varepsilon^{(j)}\right).
\end{aligned}
\]
If the errors are pairwise uncorrelated and have common variance $\sigma^2$, then all off-diagonal covariance terms vanish, giving \(\operatorname{Var}(\bar{\varepsilon}) = \frac{1}{L^2}\sum_{\ell=1}^L\sigma^2 = \frac{\sigma^2}{L}\).\\
For the median, suppose that more than half of the realized errors satisfy \(|\varepsilon^{(\ell)}|\leq\tau\). Then more than half of the errors lie in the interval $[-\tau,\tau]$. Hence, fewer than half lie strictly above $\tau$ and fewer than half lie strictly below $-\tau$. Therefore, the median cannot be outside this interval, and \(|\varepsilon_{\mathrm{med}}|\leq\tau\).
\qed

\section{Experimental Section Continued}
\label{app:nonrand}
\suppressfloats[t]

\subsection{Protocol}

We use $23$ public regression datasets, listed with their sources in
\ref{app:datasets}. Each dataset is evaluated over ten folds. Within a
fold, inputs and targets are standardised using training statistics only, the
model is fitted on the training split, and predictions are returned to the
original target scale before the test mean squared error is computed.

Hyperparameters are selected per dataset and fold by Tree-structured Parzen
Estimator search \citep{bergstra2011algorithms}, minimising cross-validated
error on the training split. The search space is given in
\ref{app:bayesian_optimization}. Depth is fixed at $\NumLayers=20$ for
all deep randomized models, and median aggregation is used throughout.

Both comparisons use the same tests. Average ranks are computed within the
family, and each baseline is tested against \LAIR by a Wilcoxon signed-rank test
over the $23$ datasets with Holm correction for multiplicity, following the
protocol recommended for comparisons over multiple datasets
\citep{demsar2006statistical} with the correction of
\citet{garcia2008extension}. The accompanying figures show the Nemenyi post-hoc
comparison, drawn against the best-ranked model, in which a model whose interval
overlaps the shaded band is not separated from it at the $0.05$ level. Testing
within a family rather than across all models at once is deliberate, and it is
what gives these tests power where it matters. Pooling all nineteen models
raises the critical difference to $5.84$. At that width the test separates \LAIR
from nine of the eighteen baselines, every one of them in the weaker half of the
field, and from none of the nine ranked immediately below it. It therefore
answers a question about the bottom of the field rather than about the
competitors this paper is arguing with.

\subsection{Datasets}
\label{app:datasets}
\suppressfloats[t]

The real-data benchmark uses the preprocessed regression files and fixed splits
from the \texttt{uci\_datasets} package. Dataset provenance is reported separately from
the package source because the package defines the exact experimental files,
whereas the dataset source identifies the data origin. Table~\ref{tab:datasets}
therefore records the local key, local shape, and provenance target for each
dataset. Several local shapes differ from the raw source description because
the benchmark uses encoded, filtered, or selected-target versions of the data.

\begin{table*}[t]
\centering
\caption{Dataset provenance targets for the benchmark suite. The local shape is the shape used by the preprocessed benchmark files.}
\label{tab:datasets}
\footnotesize
\begin{tabular}{llrrp{3.9cm}}
\toprule
Dataset & Acronym & \NumSamples & \NumFeatures & Provenance note \\
\midrule
Airfoil Self-Noise & Airfoil & 1503 & 5 & \citep{uciAirfoilSelfNoise} \\
Auto MPG & Auto MPG & 392 & 7 & \citep{uciAutoMpg}; local rows reflect preprocessing \\
Automobile & Autos & 159 & 25 & \citep{uciAutomobile}; local shape reflects encoding \\
Breast Cancer Prognostic & Breast Cancer & 194 & 33 & \citep{uciBreastCancerPrognostic} \\
Challenger O-Ring & Challenger & 23 & 4 & \citep{uciChallengerORing} \\
Concrete Strength & Concrete & 1030 & 8 & \citep{uciConcreteStrength} \\
Concrete Slump & Concrete Slump & 103 & 7 & \citep{uciConcreteSlump}; selected target \\
Energy Efficiency & Energy & 768 & 8 & \citep{uciEnergyEfficiency} \\
Fertility & Fertility & 100 & 9 & \citep{uciFertility} \\
Forest Fires & Forest Fires & 517 & 12 & \citep{uciForestFires} \\
Gas Sensor Array Drift & Gas & 2565 & 128 & \citep{uciGasSensorDrift}; regression subset \\
Boston Housing & Housing & 506 & 13 & \citep{harrison1978hedonic}; legacy dataset caveat \\
Computer Hardware & Machine CPU & 209 & 7 & \citep{uciComputerHardware}; preprocessed target and features \\
Parkinsons Telemonitoring & Parkinsons & 5875 & 20 & \citep{uciParkinsonsTelemonitoring} \\
Pendulum & Pendulum & 630 & 9 & \citep{evansUciDatasets,yang2015alacarte} \\
PumaDyn-32nm & PumaDyn-32nm & 8192 & 32 & \citep{delvePumadyn} \\
Servo & Servo & 167 & 4 & \citep{uciServo} \\
SkillCraft1 & Skillcraft & 3338 & 19 & \citep{uciSkillCraft} \\
SML2010 & SML & 4137 & 26 & \citep{uciSML2010} \\
Solar Flare & Solar & 1066 & 10 & \citep{uciSolarFlare}; corrected subset \\
Istanbul Stock Exchange & Stock & 536 & 11 & \citep{uciIstanbulStockExchange}; local feature count reflects preprocessing \\
Wine Quality & Wine & 1599 & 11 & \citep{uciWineQuality}; red-wine subset \\
Yacht Hydrodynamics & Yacht & 308 & 6 & \citep{uciYachtHydrodynamics} \\
\bottomrule
\end{tabular}
\end{table*}

The Boston Housing dataset is retained because it is part of the fixed
benchmark suite, but it is a legacy dataset and should be interpreted with that
context. The gas-sensor row is also treated cautiously. The raw UCI source is a
classification drift dataset, while the benchmark uses a preprocessed
regression subset from the package. The pendulum row remains tied to the
benchmark lineage and package source unless an authoritative original generator
is identified.

\subsection{Comparison with conventional models}

The second comparison places \LAIR against a multilayer perceptron, a decision
tree, a random forest, AdaBoost, gradient boosting, support vector regression,
$k$-nearest neighbors, and linear, ridge, lasso, and elastic-net regression.
Per-dataset errors for this family are in \ref{app:nonrand}, and the search
ranges, together with the fixed settings that constrain these baselines, are in
\ref{app:bayesian_optimization}.

\LAIR obtains the best average rank in this family as well, $3.26$ of twelve
models, ahead of the multilayer perceptron at $3.87$ and the random forest at
$4.96$. Every baseline is rejected after Holm correction, with the largest
adjusted $p$-value $0.0389$.

\begin{figure}[t]
\centering
\includegraphics[width=0.88\columnwidth]{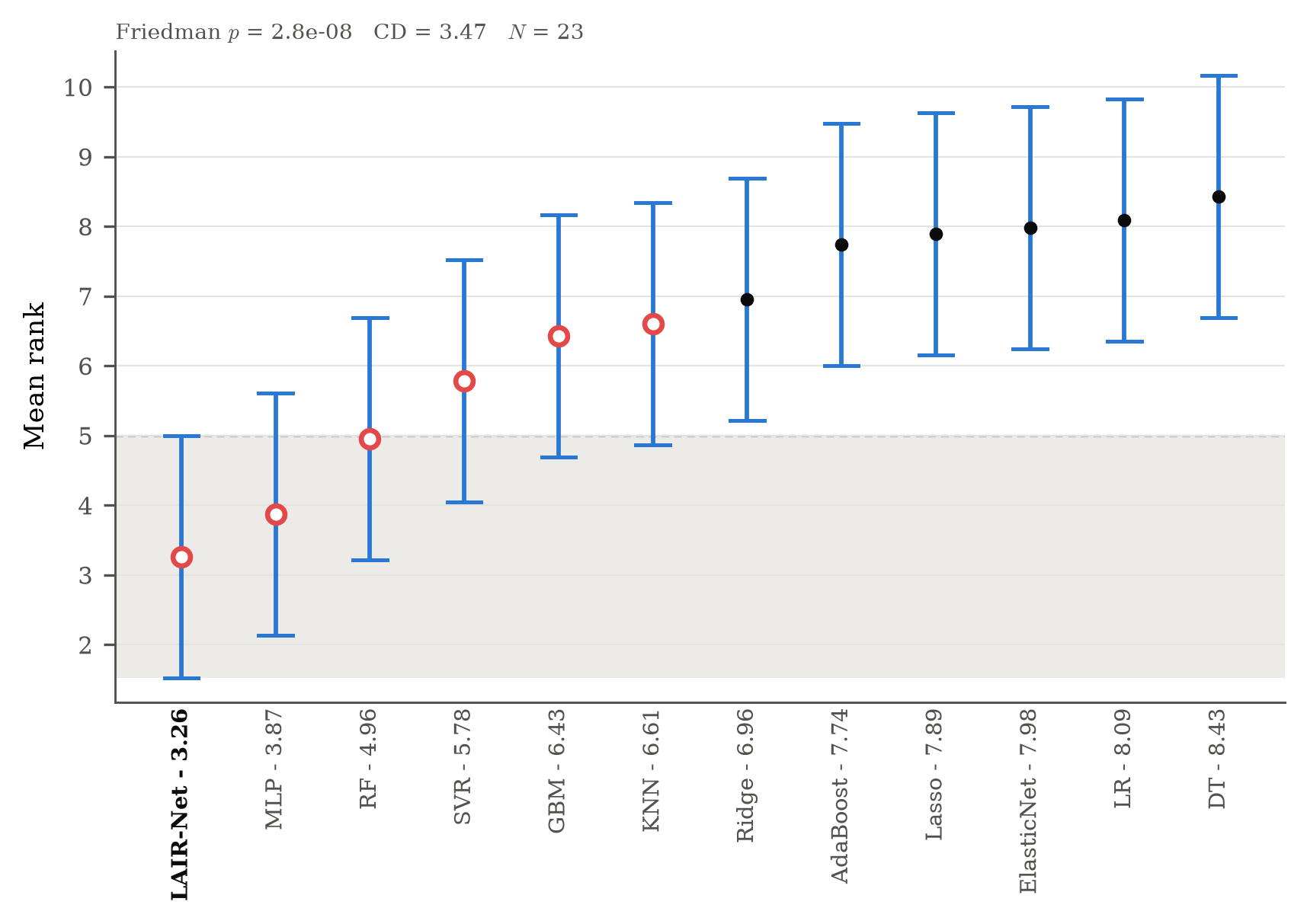}
\caption{Nemenyi comparison against conventional models, drawn as in
Figure~\ref{fig:nemenyi_rand}, with $\mathrm{CD}=3.47$ over $23$ datasets.
Lower is better.}
\label{fig:nemenyi_nonrand}
\end{figure}

Figure~\ref{fig:nemenyi_nonrand} is the more cautious of the two. The wider
critical difference, $3.47$ against twelve models rather than $2.19$ against
eight, separates \LAIR from six of the eleven baselines and leaves it
unseparated from the multilayer perceptron, the random forest, support vector
regression, gradient boosting and $k$-nearest neighbours, even though it
outranks all five, and the pairwise tests reject them. What this comparison
supports is that \LAIR is competitive outside its own family on these datasets,
not that it displaces the methods in it. That distinction matters on the tabular
data, where carefully tuned classical methods remain competitive with deep
architectures \citep{grinsztajn2022why,shwartz2022tabular}.

Section~\ref{sec:experiments} reports the comparison against conventional
models through its average ranks, its Holm-adjusted tests and
Figure~\ref{fig:nemenyi_nonrand}. Table~\ref{tab:benchmark_nonrand} gives the
per-dataset errors behind those summaries. Search ranges for these models are in
Table~\ref{tab:searchspace_nonrand}, and the fixed settings that constrain them
are stated alongside it.

\begin{table*}[htbp]
\centering
\caption{Test mean squared error against conventional neural, tree-based, kernel, instance-based and linear models, mean $\pm$ standard deviation over ten folds. Lower is better and the best entry in each row is in bold. The final rows give the average rank within this family and the Holm-adjusted Wilcoxon signed-rank $p$-value against \LAIR.}
\label{tab:benchmark_nonrand}
\resizebox{\textwidth}{!}{%
\begin{tabular}{lcccccccccccc}
\toprule
Dataset & MLP & DT & RF & AdaBoost & GBM & SVR & KNN & LR & Ridge & Lasso & ElasticNet & LAIR-Net \\
\midrule
Airfoil & 5.315 \tiny{$\pm$ 1.512} & 10.705 \tiny{$\pm$ 2.487} & 7.501 \tiny{$\pm$ 1.896} & 21.337 \tiny{$\pm$ 4.179} & 16.551 \tiny{$\pm$ 2.630} & 9.192 \tiny{$\pm$ 1.916} & 3.868 \tiny{$\pm$ 0.786} & 23.378 \tiny{$\pm$ 5.534} & 23.378 \tiny{$\pm$ 5.532} & 23.378 \tiny{$\pm$ 5.532} & 23.375 \tiny{$\pm$ 5.530} & \textbf{2.208 \tiny{$\pm$ 0.485}} \\
Auto MPG & 7.483 \tiny{$\pm$ 1.514} & 12.560 \tiny{$\pm$ 2.468} & 8.538 \tiny{$\pm$ 2.757} & 11.207 \tiny{$\pm$ 3.339} & 9.825 \tiny{$\pm$ 2.377} & \textbf{6.734 \tiny{$\pm$ 1.975}} & 7.880 \tiny{$\pm$ 2.700} & 11.266 \tiny{$\pm$ 2.480} & 11.252 \tiny{$\pm$ 2.453} & 11.399 \tiny{$\pm$ 2.619} & 11.400 \tiny{$\pm$ 2.525} & 6.919 \tiny{$\pm$ 2.074} \\
Autos & 0.027 \tiny{$\pm$ 0.013} & 0.034 \tiny{$\pm$ 0.011} & 0.029 \tiny{$\pm$ 0.015} & 0.031 \tiny{$\pm$ 0.015} & 0.027 \tiny{$\pm$ 0.018} & 0.028 \tiny{$\pm$ 0.010} & 0.031 \tiny{$\pm$ 0.022} & 0.027 \tiny{$\pm$ 0.010} & 0.024 \tiny{$\pm$ 0.007} & 0.027 \tiny{$\pm$ 0.013} & 0.027 \tiny{$\pm$ 0.012} & \textbf{0.021 \tiny{$\pm$ 0.010}} \\
Breast Cancer & 939.197 \tiny{$\pm$ 333.499} & 1.06e3 \tiny{$\pm$ 253.065} & 832.325 \tiny{$\pm$ 208.431} & 839.905 \tiny{$\pm$ 229.339} & \textbf{741.678 \tiny{$\pm$ 162.746}} & 911.205 \tiny{$\pm$ 290.348} & 1.02e3 \tiny{$\pm$ 236.385} & 944.958 \tiny{$\pm$ 364.476} & 896.263 \tiny{$\pm$ 308.580} & 914.506 \tiny{$\pm$ 276.342} & 874.418 \tiny{$\pm$ 257.062} & 871.580 \tiny{$\pm$ 265.293} \\
Challenger & 0.375 \tiny{$\pm$ 0.436} & 0.407 \tiny{$\pm$ 0.365} & 0.407 \tiny{$\pm$ 0.370} & 0.480 \tiny{$\pm$ 0.556} & 0.415 \tiny{$\pm$ 0.358} & \textbf{0.317 \tiny{$\pm$ 0.314}} & 0.428 \tiny{$\pm$ 0.397} & 0.365 \tiny{$\pm$ 0.408} & 0.359 \tiny{$\pm$ 0.391} & 0.414 \tiny{$\pm$ 0.451} & 0.395 \tiny{$\pm$ 0.432} & 0.389 \tiny{$\pm$ 0.448} \\
Concrete & 43.674 \tiny{$\pm$ 12.915} & 74.354 \tiny{$\pm$ 26.260} & 46.642 \tiny{$\pm$ 9.922} & 79.306 \tiny{$\pm$ 15.171} & 65.111 \tiny{$\pm$ 10.004} & 42.791 \tiny{$\pm$ 9.954} & 58.822 \tiny{$\pm$ 11.054} & 110.468 \tiny{$\pm$ 12.515} & 110.461 \tiny{$\pm$ 12.455} & 110.905 \tiny{$\pm$ 11.686} & 110.962 \tiny{$\pm$ 11.709} & \textbf{40.199 \tiny{$\pm$ 10.883}} \\
Concrete Slump & 76.849 \tiny{$\pm$ 46.411} & 3.69e3 \tiny{$\pm$ 1.26e3} & 2.27e3 \tiny{$\pm$ 782.759} & 2.16e3 \tiny{$\pm$ 1.14e3} & 2.64e3 \tiny{$\pm$ 1.42e3} & 66.261 \tiny{$\pm$ 33.492} & 2.82e3 \tiny{$\pm$ 1.45e3} & 63.754 \tiny{$\pm$ 40.893} & 63.703 \tiny{$\pm$ 42.160} & 71.266 \tiny{$\pm$ 50.263} & 71.815 \tiny{$\pm$ 50.085} & \textbf{11.170 \tiny{$\pm$ 10.246}} \\
Energy & 0.397 \tiny{$\pm$ 0.138} & 1.024 \tiny{$\pm$ 0.749} & 1.055 \tiny{$\pm$ 0.730} & 5.107 \tiny{$\pm$ 1.681} & 4.565 \tiny{$\pm$ 0.914} & 3.177 \tiny{$\pm$ 0.813} & 4.273 \tiny{$\pm$ 1.187} & 8.172 \tiny{$\pm$ 1.861} & 8.730 \tiny{$\pm$ 1.899} & 8.748 \tiny{$\pm$ 1.909} & 8.735 \tiny{$\pm$ 1.900} & \textbf{0.163 \tiny{$\pm$ 0.045}} \\
Fertility & 0.033 \tiny{$\pm$ 0.016} & 0.032 \tiny{$\pm$ 0.020} & \textbf{0.029 \tiny{$\pm$ 0.018}} & 0.032 \tiny{$\pm$ 0.022} & 0.031 \tiny{$\pm$ 0.016} & 0.036 \tiny{$\pm$ 0.018} & 0.034 \tiny{$\pm$ 0.012} & 0.032 \tiny{$\pm$ 0.013} & 0.032 \tiny{$\pm$ 0.013} & 0.031 \tiny{$\pm$ 0.012} & 0.032 \tiny{$\pm$ 0.013} & 0.033 \tiny{$\pm$ 0.019} \\
Forest Fires & \textbf{1.943 \tiny{$\pm$ 0.442}} & 2.005 \tiny{$\pm$ 0.433} & 1.966 \tiny{$\pm$ 0.448} & 1.990 \tiny{$\pm$ 0.507} & 1.982 \tiny{$\pm$ 0.490} & 1.961 \tiny{$\pm$ 0.421} & 1.985 \tiny{$\pm$ 0.441} & 2.094 \tiny{$\pm$ 0.625} & 2.088 \tiny{$\pm$ 0.614} & 1.990 \tiny{$\pm$ 0.457} & 1.991 \tiny{$\pm$ 0.456} & 2.073 \tiny{$\pm$ 0.582} \\
Gas & 0.042 \tiny{$\pm$ 0.029} & 0.054 \tiny{$\pm$ 0.041} & 0.034 \tiny{$\pm$ 0.020} & 0.097 \tiny{$\pm$ 0.026} & 0.066 \tiny{$\pm$ 0.019} & 0.066 \tiny{$\pm$ 0.059} & \textbf{0.025 \tiny{$\pm$ 0.025}} & 1.04e5 \tiny{$\pm$ 3.29e5} & 317.597 \tiny{$\pm$ 1.00e3} & 0.483 \tiny{$\pm$ 0.908} & 0.358 \tiny{$\pm$ 0.406} & 8.752 \tiny{$\pm$ 27.367} \\
Housing & 12.249 \tiny{$\pm$ 8.279} & 20.344 \tiny{$\pm$ 15.080} & 16.036 \tiny{$\pm$ 12.688} & 20.077 \tiny{$\pm$ 13.640} & 16.246 \tiny{$\pm$ 11.726} & 11.370 \tiny{$\pm$ 8.452} & 15.945 \tiny{$\pm$ 14.277} & 24.072 \tiny{$\pm$ 10.927} & 24.216 \tiny{$\pm$ 11.144} & 24.057 \tiny{$\pm$ 10.958} & 24.101 \tiny{$\pm$ 10.973} & \textbf{10.706 \tiny{$\pm$ 8.377}} \\
Machine CPU & 0.181 \tiny{$\pm$ 0.040} & 0.276 \tiny{$\pm$ 0.084} & 0.194 \tiny{$\pm$ 0.047} & 0.202 \tiny{$\pm$ 0.055} & 0.192 \tiny{$\pm$ 0.053} & 0.242 \tiny{$\pm$ 0.072} & 0.181 \tiny{$\pm$ 0.067} & 0.226 \tiny{$\pm$ 0.048} & 0.226 \tiny{$\pm$ 0.050} & 0.226 \tiny{$\pm$ 0.046} & 0.227 \tiny{$\pm$ 0.046} & \textbf{0.146 \tiny{$\pm$ 0.034}} \\
Parkinsons & 78.795 \tiny{$\pm$ 23.170} & 82.672 \tiny{$\pm$ 29.929} & 45.005 \tiny{$\pm$ 20.361} & 47.374 \tiny{$\pm$ 5.028} & 61.041 \tiny{$\pm$ 15.753} & 107.940 \tiny{$\pm$ 30.178} & 27.980 \tiny{$\pm$ 2.585} & 86.108 \tiny{$\pm$ 2.380} & 86.022 \tiny{$\pm$ 2.328} & 109.409 \tiny{$\pm$ 10.549} & 97.089 \tiny{$\pm$ 1.883} & \textbf{24.092 \tiny{$\pm$ 17.750}} \\
Pendulum & 2.393 \tiny{$\pm$ 1.438} & 6.809 \tiny{$\pm$ 2.977} & 5.399 \tiny{$\pm$ 2.394} & 8.277 \tiny{$\pm$ 3.890} & 6.158 \tiny{$\pm$ 2.700} & 3.910 \tiny{$\pm$ 2.729} & 3.768 \tiny{$\pm$ 2.267} & 7.495 \tiny{$\pm$ 2.875} & 7.488 \tiny{$\pm$ 2.895} & 7.458 \tiny{$\pm$ 2.948} & 7.471 \tiny{$\pm$ 2.949} & \textbf{1.681 \tiny{$\pm$ 0.963}} \\
PumaDyn-32nm & 0.089 \tiny{$\pm$ 0.006} & 0.135 \tiny{$\pm$ 0.020} & 0.111 \tiny{$\pm$ 0.009} & 0.816 \tiny{$\pm$ 0.101} & 0.821 \tiny{$\pm$ 0.142} & 0.752 \tiny{$\pm$ 0.037} & 0.950 \tiny{$\pm$ 0.068} & 1.002 \tiny{$\pm$ 0.065} & 1.002 \tiny{$\pm$ 0.065} & 0.995 \tiny{$\pm$ 0.066} & 0.995 \tiny{$\pm$ 0.066} & \textbf{0.052 \tiny{$\pm$ 0.004}} \\
Servo & \textbf{0.084 \tiny{$\pm$ 0.054}} & 0.128 \tiny{$\pm$ 0.058} & 0.153 \tiny{$\pm$ 0.066} & 0.124 \tiny{$\pm$ 0.042} & 0.127 \tiny{$\pm$ 0.058} & 0.088 \tiny{$\pm$ 0.062} & 0.090 \tiny{$\pm$ 0.057} & 0.333 \tiny{$\pm$ 0.127} & 0.332 \tiny{$\pm$ 0.127} & 0.332 \tiny{$\pm$ 0.127} & 0.333 \tiny{$\pm$ 0.127} & 0.086 \tiny{$\pm$ 0.047} \\
SkillCraft & \textbf{0.063 \tiny{$\pm$ 0.009}} & 0.077 \tiny{$\pm$ 0.008} & 0.070 \tiny{$\pm$ 0.011} & 0.078 \tiny{$\pm$ 0.010} & 0.074 \tiny{$\pm$ 0.009} & 0.083 \tiny{$\pm$ 0.060} & 0.076 \tiny{$\pm$ 0.009} & 0.064 \tiny{$\pm$ 0.009} & 0.064 \tiny{$\pm$ 0.009} & 0.064 \tiny{$\pm$ 0.009} & 0.064 \tiny{$\pm$ 0.009} & 0.066 \tiny{$\pm$ 0.011} \\
SML & 0.331 \tiny{$\pm$ 0.074} & 0.936 \tiny{$\pm$ 0.165} & 0.497 \tiny{$\pm$ 0.065} & 4.480 \tiny{$\pm$ 0.506} & 3.855 \tiny{$\pm$ 0.350} & 0.585 \tiny{$\pm$ 0.136} & 0.837 \tiny{$\pm$ 0.125} & 2.358 \tiny{$\pm$ 0.198} & 2.358 \tiny{$\pm$ 0.198} & 2.402 \tiny{$\pm$ 0.193} & 2.359 \tiny{$\pm$ 0.196} & \textbf{0.117 \tiny{$\pm$ 0.023}} \\
Solar & 0.623 \tiny{$\pm$ 0.353} & 0.637 \tiny{$\pm$ 0.363} & 0.604 \tiny{$\pm$ 0.352} & 0.620 \tiny{$\pm$ 0.362} & \textbf{0.599 \tiny{$\pm$ 0.352}} & 0.697 \tiny{$\pm$ 0.402} & 0.622 \tiny{$\pm$ 0.341} & 0.617 \tiny{$\pm$ 0.342} & 0.617 \tiny{$\pm$ 0.343} & 0.616 \tiny{$\pm$ 0.346} & 0.616 \tiny{$\pm$ 0.347} & 0.622 \tiny{$\pm$ 0.348} \\
Stock & 2.57e-5 \tiny{$\pm$ 5.09e-6} & 5.42e-5 \tiny{$\pm$ 1.52e-5} & 3.81e-5 \tiny{$\pm$ 1.22e-5} & 3.95e-5 \tiny{$\pm$ 1.09e-5} & 4.41e-5 \tiny{$\pm$ 1.42e-5} & 2.57e-5 \tiny{$\pm$ 4.85e-6} & 8.70e-5 \tiny{$\pm$ 1.96e-5} & \textbf{2.53e-5 \tiny{$\pm$ 5.03e-6}} & \textbf{2.53e-5 \tiny{$\pm$ 5.01e-6}} & \textbf{2.53e-5 \tiny{$\pm$ 4.95e-6}} & 2.53e-5 \tiny{$\pm$ 4.98e-6} & 2.56e-5 \tiny{$\pm$ 3.88e-6} \\
Wine Quality & 0.234 \tiny{$\pm$ 0.038} & 0.426 \tiny{$\pm$ 0.083} & 0.313 \tiny{$\pm$ 0.079} & 0.527 \tiny{$\pm$ 0.075} & 0.449 \tiny{$\pm$ 0.070} & 0.248 \tiny{$\pm$ 0.043} & 0.309 \tiny{$\pm$ 0.053} & 0.357 \tiny{$\pm$ 0.047} & 0.357 \tiny{$\pm$ 0.047} & 0.357 \tiny{$\pm$ 0.047} & 0.357 \tiny{$\pm$ 0.047} & \textbf{0.218 \tiny{$\pm$ 0.038}} \\
Yacht & 0.075 \tiny{$\pm$ 0.070} & 0.216 \tiny{$\pm$ 0.159} & 0.125 \tiny{$\pm$ 0.087} & 0.104 \tiny{$\pm$ 0.061} & 0.257 \tiny{$\pm$ 0.191} & 0.107 \tiny{$\pm$ 0.093} & 0.258 \tiny{$\pm$ 0.217} & 0.104 \tiny{$\pm$ 0.094} & 0.104 \tiny{$\pm$ 0.094} & 0.104 \tiny{$\pm$ 0.094} & 0.104 \tiny{$\pm$ 0.093} & \textbf{0.020 \tiny{$\pm$ 0.032}} \\
\midrule
Average rank & 3.87 & 8.43 & 4.96 & 7.74 & 6.43 & 5.78 & 6.61 & 8.09 & 6.96 & 7.87 & 7.96 & 3.26 \\
Holm-adjusted $p$ & 0.0389 & 0.0078 & 0.0389 & 0.0389 & 0.0389 & 0.0389 & 0.0078 & 0.0027 & 0.0028 & 0.0115 & 0.0115 & -- \\
\bottomrule
\end{tabular}}
\end{table*}

\begin{table}[t]
\centering
\caption{Average rank and paired Wilcoxon signed-rank tests against \LAIR{} with all nineteen models pooled, over the $23$ datasets. Lower rank is better. $p$-values are Holm-adjusted over the eighteen comparisons. Ranks here are not comparable with the within-family tables, since each model is ranked against a different set.}
\label{tab:ranktest_all}
\begin{tabular}{lccc}
\toprule
Model & Avg.\ rank & $p$ & $p_{\mathrm{Holm}}$ \\
\midrule
\textbf{LAIR-Net} & \textbf{4.22} & -- & -- \\
\midrule
MLP & 6.70 & 0.0078 & 0.0389 \\
edRVFL-SC & 7.17 & 0.0037 & 0.0221 \\
RVFL-Boost & 7.35 & 0.0010 & 0.0106 \\
RVFL-Bag & 8.00 & 0.0022 & 0.0157 \\
edRVFL & 8.35 & 0.0001 & 0.0022 \\
dRVFL & 8.52 & $<10^{-4}$ & 0.0008 \\
RF & 8.57 & 0.0217 & 0.0389 \\
SVR & 9.48 & 0.0078 & 0.0389 \\
RVFL & 9.87 & 0.0004 & 0.0046 \\
GBM & 10.91 & 0.0132 & 0.0389 \\
KNN & 10.91 & 0.0010 & 0.0106 \\
Ridge & 12.09 & 0.0003 & 0.0039 \\
Fuzzy RVFL & 12.39 & $<10^{-4}$ & 0.0008 \\
AdaBoost & 12.43 & 0.0085 & 0.0389 \\
ElasticNet & 12.83 & 0.0018 & 0.0148 \\
Lasso & 12.91 & 0.0016 & 0.0148 \\
DT & 13.57 & 0.0009 & 0.0104 \\
LR & 13.70 & 0.0002 & 0.0037 \\
\bottomrule
\end{tabular}
\end{table}

\section{Ablation and mechanism-variant study}
\label{app:ablation}
\suppressfloats[t]

This study compares the complete \LAIR construction with six controlled
variants. Three variants remove or replace the supervised alignment impulse,
one changes the initial state, and two change the random-weight and readout
mechanisms. The last two are therefore mechanism variants rather than strict
ablations. This distinction matters: a performance difference involving those
models cannot be attributed to a single removed component.

\subsection{Reference configuration: \LAIR}

All comparisons begin with the same input-recursive residual construction. At
depth $\LayerIdx$, the random feature block receives the bias, the original
input, and the preceding hidden state, as defined in
\eqref{eq:layer-input}. It generates $\RandHidden{\LayerIdx}$ through
\eqref{eq:rand-response}, after which the leaky state is

\begin{equation}
\LeakState{\LayerIdx}
=
\LeakCoef\RandHidden{\LayerIdx}
+
(1-\LeakCoef)\HiddenState{\LayerIdx-1}.
\label{eq:ablation-leak}
\end{equation}

The complete \LAIR model starts from $\HiddenState{0}=\mathbf{0}$ and injects
the target-aware \SLAN anchor at every depth:

\begin{equation}
\HiddenState{\LayerIdx}
=
(1-\AlignCoef)\LeakState{\LayerIdx}
+
\AlignCoef\AnchorMat.
\label{eq:ablation-reference}
\end{equation}

Its hidden weights are sampled independently from a scaled uniform
distribution and then fixed. Each layer uses the standard ridge readout in
\eqref{eq:ridge}, and the final prediction is the elementwise median of the
layerwise predictions. Consequently, a variant differs from \LAIR only in the
components stated below; input recursion, layerwise readouts, and median
aggregation remain unchanged unless explicitly noted.

\subsection{Alignment ablations}

\paragraph{\LAIRNoAlign: removal of supervised alignment.}\hspace{0.4em}
This is the strict no-alignment ablation. The \SLAN is not fitted, no anchor is
constructed, and $\AlignCoef$ is effectively zero. The state update is simply

\begin{equation}
\HiddenState{\LayerIdx}=\LeakState{\LayerIdx},
\qquad
\HiddenState{0}=\mathbf{0}.
\label{eq:ablation-no-align}
\end{equation}

The comparison with \LAIR isolates the contribution of the target-aware
alignment pathway while retaining the leaky residual recursion, direct input
access at every depth, the same type of random hidden weights, layerwise ridge
readouts, and median aggregation. It therefore asks whether leakage and input
recursion alone are sufficient, or whether repeatedly returning the state
towards a supervised representation improves cross-dataset consistency.

\paragraph{\LAIRRandomImpulse: target-independent fixed impulse.}\hspace{0.4em}
This variant replaces the supervised anchor with a random feature map. A
single random matrix $\mathbf{W}_{\mathrm{c}}$ is sampled before the depth-wise
recursion and used to construct

\begin{equation}
\mathbf{C}
=
\Activation\!\left(\AugInput\mathbf{W}_{\mathrm{c}}\right).
\label{eq:ablation-random-anchor}
\end{equation}

The same matrix $\mathbf{C}$ is injected at every layer:

\begin{equation}
\HiddenState{\LayerIdx}
=
(1-\AlignCoef)\LeakState{\LayerIdx}
+
\AlignCoef\mathbf{C},
\qquad
\HiddenState{0}=\mathbf{0}.
\label{eq:ablation-random-impulse}
\end{equation}

The word ``fixed'' refers to depth: $\mathbf{C}$ is recomputed from each
sample's covariates but does not change from one randomised layer to the next.
Unlike $\AnchorMat$, it is independent of the response values and receives no
supervised training. The comparison with \LAIR separates the effect of having
a persistent reference from the effect of making that reference target-aware.

\paragraph{\LAIRBestState: data-selected recurrent impulse.}\hspace{0.4em}
This variant removes the \SLAN and uses an earlier hidden state as the impulse.
After fitting the readout at each layer, the implementation evaluates its
training leave-one-out cross-validation (LOOCV) error. The hidden state with
the smallest LOOCV error observed so far is retained as
$\mathbf{H}_{\mathrm{best}}$. For layers after the first, the update is

\begin{equation}
\HiddenState{\LayerIdx}
=
(1-\AlignCoef)\LeakState{\LayerIdx}
+
\AlignCoef\mathbf{H}_{\mathrm{best}}^{(\LayerIdx-1)}.
\label{eq:ablation-best-state}
\end{equation}

No previous state exists at the first layer, so the implementation uses
$\LeakState{1}$ as the impulse and obtains
$\HiddenState{1}=\LeakState{1}$. During prediction, the sequence of best-state
updates is replayed from the indices selected on the training data; test
targets are never used. This comparison asks whether adaptive reuse of a
previously successful random state can replace the external target-aware
reference supplied by the \SLAN.

\subsection{Initial-state and orthogonal-weight variants}

\paragraph{\LAIRSLANInit (\SLAN-initialised LAIR)}
This model differs from \LAIR only in the initial condition:

\begin{equation}
\HiddenState{0}=\AnchorMat.
\label{eq:ablation-slan-init}
\end{equation}

For every $\LayerIdx\geq 1$, it still applies the complete update in
\eqref{eq:ablation-reference}; hence, the \SLAN anchor is both the initial state
and a repeated alignment impulse. Because all subsequent alignment impulses are retained,
the comparison between \LAIR and \LAIRSLANInit isolates the initial condition,
not the value of repeated alignment.

\paragraph{\LAIROrtho (orthogonal random transformations)}
This variant retains the \SLAN, starts from
$\HiddenState{0}=\AnchorMat$, and injects $\AnchorMat$ at every depth using
\eqref{eq:ablation-reference}. Its principal additional change is the
construction of each random hidden-weight matrix. A square Gaussian matrix is
factorised by QR decomposition, an appropriately sized block of its orthogonal
factor is retained, and that block is multiplied by the input-scale
hyperparameter. In contrast, \LAIR samples independent uniform entries.

The layer input remains $[\BiasVec,\InputMat,\HiddenState{\LayerIdx-1}]$ in
both models. Thus, input recursion is a shared property and should not be
presented as the distinguishing feature of \LAIROrtho. Standard ridge readouts
and median aggregation are also retained. Since this model changes both the
initial state and the random-weight geometry, its comparison with \LAIR does
not isolate orthogonality alone; it evaluates the combined orthogonal,
\SLAN-initialised configuration.

\paragraph{\LAIROrthoBMA: orthogonal model with LOOCV-weighted aggregation.}\hspace{0.4em}
This variant uses the same orthogonal hidden weights, \SLAN initialisation,
repeated alignment impulse, and input-recursive layer construction as
\LAIROrtho. It then changes both readout regularisation and prediction
aggregation. For a design matrix
$\DesignMat{\LayerIdx}=[\BiasVec,\InputMat,\HiddenState{\LayerIdx}]$, the bias
and original-input coefficients receive penalty
$\RidgeReg_{\mathrm{x}}=r\RidgeReg$, whereas the hidden-state coefficients
receive $\RidgeReg_{\mathrm{h}}=\RidgeReg$. The ratio $r$ is tuned separately.

Let $e_{\LayerIdx}$ denote the training LOOCV error of the layer-$\LayerIdx$
readout and let $s_e$ denote the standard deviation of the negative LOOCV
scores. The aggregation weights are

\begin{equation}
\omega_{\LayerIdx}
=
\frac{\exp\!\left[-e_{\LayerIdx}/(s_e\tau)\right]}
{\sum_{j=1}^{\NumLayers}
 \exp\!\left[-e_j/(s_e\tau)\right]},
\qquad
\FinalPred
=
\sum_{\LayerIdx=1}^{\NumLayers}
\omega_{\LayerIdx}\LayerPred{\LayerIdx},
\label{eq:ablation-bma}
\end{equation}

where $\tau>0$ is a temperature hyperparameter. Lower-LOOCV-error layers
therefore receive greater weight, whereas \LAIR and \LAIROrtho use an
unweighted elementwise median. The comparison between \LAIROrtho and
\LAIROrthoBMA evaluates the joint effect of dual-penalty ridge estimation and
LOOCV-weighted aggregation; it cannot distinguish those two changes from each
other.

\subsection{Results within the variant family}

Table~\ref{tab:ablation_summary} reports the average rank and number of
per-dataset wins for each model within the variant family. The two measures
capture complementary aspects of performance. \LAIR achieves the lowest
average rank of $2.565$, indicating the strongest overall performance across
the benchmark suite. In contrast, \LAIRSLANInit records the largest number of
dataset wins and the second-lowest average rank. Thus, although the
\SLAN-initialised configuration performs best on more individual datasets,
the zero-initialised \LAIR configuration is more consistently competitive
across the full suite.

\begin{table}[b]
\centering
\caption{Average ranks and per-dataset wins within the variant family.
Lower average rank and higher win count are better.}
\label{tab:ablation_summary}
\begin{tabular}{lcc}
\toprule
Model & Average rank & Dataset wins \\
\midrule
\LAIR                 & \textbf{2.565} & 3 \\
\LAIRSLANInit         & 3.043          & \textbf{7} \\
\LAIROrtho            & 3.783          & 4 \\
\LAIROrthoBMA         & 3.913          & 4 \\
\LAIRNoAlign          & 4.652          & 3 \\
\LAIRBestState        & 4.696          & 2 \\
\LAIRRandomImpulse    & 5.348          & 0 \\
\bottomrule
\end{tabular}
\end{table}

Most datasets where \LAIR is not the best-performing model remain close to the best
variant. The gas-sensor dataset is the exception: \LAIR ranks second within the
variant family but has a much larger \(\MSE\) than \LAIRSLANInit. This dataset
is already treated separately in the real-data analysis, where its first-layer
standardisation behaviour is atypical. It should therefore be interpreted as
evidence that the initial condition can matter strongly on a particular
dataset, rather than as evidence against repeated alignment: both models inject
the same \SLAN anchor at every depth.

Taken together, the results support three bounded conclusions. First, the
complete \LAIR configuration has the best average rank in this comparison.
Second, replacing the supervised anchor by a target-independent random impulse
or by a running-best hidden state does not reproduce that average-rank
performance. Third, changing the initial state to $\AnchorMat$ increases the
number of dataset wins but reduces average-rank consistency. The orthogonal
variants remain competitive, but because they alter multiple mechanisms they
should be interpreted as alternative configurations rather than evidence for
or against any single \LAIR component.

\section{Bayesian optimisation and search space}
\label{app:bayesian_optimization}
\suppressfloats[t]

Validation error is an expensive black-box function of the hyperparameter
vector, so hyperparameters are selected by the tree-structured Parzen estimator
\citep{bergstra2011algorithms} rather than by grid or random search. Instead of
modelling the objective directly, the estimator models the distribution of
hyperparameters associated with good and with poor observed values, and proposes
the configuration that maximises the ratio of the two densities. This suits a
search space that mixes continuous, integer and categorical parameters, which is
the case here.

For each dataset and outer fold, the search minimises five-fold
cross-validated mean squared error on the training split. Inputs and targets
are standardised using training-fold statistics only. Predictions are returned
to the original scale before the error is computed. The configuration with the
lowest objective is then refitted on the full training split before the held-out
test error is recorded. The test split is not used at any point in the search.

\subsection{Searched hyperparameters}

Tables~\ref{tab:searchspace} and~\ref{tab:searchspace_nonrand} list the searched
ranges for the two comparison families. Continuous ranges written
$\log\mathcal{U}$ are sampled uniformly in the logarithm, which is appropriate
for scale parameters spanning several orders of magnitude.

\begin{table}[htbp]
\centering
\caption{Hyperparameter search spaces for the randomized models.}
\label{tab:searchspace}
\begin{tabular}{@{}ll@{}}
\toprule
Hyperparameter & Search space \\
\midrule

\multicolumn{2}{@{}l}{\textit{Shared by all randomized models, including \LAIR}} \\[2pt]
\quad $\HiddenWidth$  & $\{10,20,\ldots,200\}$ \\
\quad Input scale     & $\mathcal{U}(0,1)$ \\
\quad $\RidgeReg$     & $\log\mathcal{U}(10^{-8},10^{2})$ \\
\quad Activation      & $\{\mathrm{ReLU},\tanh\}$ \\

\addlinespace
\multicolumn{2}{@{}l}{\textit{\LAIR-specific}} \\[2pt]
\quad $\LeakCoef$     & $\mathcal{U}(0,1)$ \\
\quad $\AlignCoef$    & $\mathcal{U}(0,1)$ \\
\quad $\AlignReg$     & $\log\mathcal{U}(10^{-8},10^{2})$ \\

\addlinespace
\multicolumn{2}{@{}l}{\textit{Fuzzy \RVFL-specific}} \\[2pt]
\quad Number of rules & $\{20,21,\ldots,40\}$ \\

\bottomrule
\end{tabular}
\end{table}

The conventional models of Table~\ref{tab:benchmark_nonrand} are searched by the
same estimator, over the same number of trials, under the same nested protocol.
Their ranges are given in Table~\ref{tab:searchspace_nonrand}.

\begin{table}[htbp]
\centering
\caption{Hyperparameter search space for the conventional models. Ridge and
lasso regression are fitted by their own internal cross-validation over the
regularisation path rather than by the outer search, and linear regression has
no hyperparameter, so none of the three enters this table.}
\label{tab:searchspace_nonrand}
\small
\begin{tabular}{lll}
\toprule
Model & Parameter & Range \\
\midrule
MLP & hidden units & $\{8,16,32,64\}$ \\
    & activation & $\{\text{identity},\text{logistic},\text{tanh},\text{ReLU}\}$ \\
    & learning-rate schedule & $\{\text{constant},\text{invscaling},\text{adaptive}\}$ \\
    & initial learning rate & $\{10^{-1},10^{-2},10^{-3},10^{-4}\}$ \\
\midrule
Decision tree & criterion & $\{\text{squared},\text{absolute},\text{Friedman}\}$ \\
              & splitter & $\{\text{best},\text{random}\}$ \\
              & min.\ samples per leaf & $\{8,32,64,96,128\}$ \\
              & max.\ features & $\{0.2,0.4,0.8,\sqrt{\cdot},\log_2\}$ \\
\midrule
Random forest & min.\ samples per leaf & $\{8,32,64,96,128\}$ \\
              & max.\ features & $\{0.2,0.4,0.8,\sqrt{\cdot},\log_2\}$ \\
\midrule
AdaBoost & learning rate & $\{10^{-1},10^{-2},10^{-3}\}$ \\
         & loss & $\{\text{linear},\text{square},\text{exponential}\}$ \\
\midrule
Gradient boosting & learning rate & $\{10^{-1},10^{-2},10^{-3}\}$ \\
                  & min.\ samples per leaf & $\{8,32,128\}$ \\
                  & max.\ features & $\{0.2,0.4,0.8,\sqrt{\cdot},\log_2\}$ \\
\midrule
SVR & $C$ & $\mathcal{U}(0.01,3)$ \\
    & $\epsilon$ & $\mathcal{U}(0.1,1)$ \\
    & kernel & $\{\text{linear},\text{polynomial},\text{RBF}\}$ \\
    & polynomial degree & $\{1,2,3,4,5\}$ \\
\midrule
$k$-nearest neighbours & $k$ & $\{1,2,\dots,50\}$ \\
                       & weighting & $\{\text{uniform},\text{distance}\}$ \\
\midrule
Elastic net & $\alpha$ & $\mathcal{U}(0,3)$ \\
            & $\ell_1$ ratio & $\mathcal{U}(0,1)$ \\
\bottomrule
\end{tabular}
\end{table}

The following are held fixed rather than searched, so that models are compared
at matched depth and matched aggregation. The number of layers is
$\NumLayers=20$ for all deep randomized models, including \dRVFL, \edRVFL,
\edRVFLSC and \LAIR. Layerwise predictions are combined by the median. The
alignment network is optimised by L-BFGS \citep{liu1989limited} with an
iteration cap of $2000$, and its activation is matched to the activation of the
randomized layers rather than searched separately. The random seed for weight
generation is fixed across models within a fold, so that a difference between
two models is not a difference in draws.

Three fixed settings on the conventional side are worth stating plainly, because
each of them constrains a baseline. The random forest, AdaBoost and gradient
boosting all use $20$ estimators, which is modest for those methods and is not
searched. Support vector regression is capped at $1000$ iterations. Ridge and
lasso use their internal cross-validated variants rather than entering the outer
search. A reader should read the conventional comparison with those limits in
view, and we would expect the tree ensembles in particular to improve with more
estimators.

The ranges above are those implemented in the accompanying code, and the study
reported in Section~\ref{sec:experiments} uses them without modification. Two
consequences are worth recording for anyone reproducing the results. First, the
lower end of the ridge range reaches $10^{-8}$, so configurations near that end
can produce fitted readouts that are sensitive to the linear algebra
implementation. Second, the alignment network's L-BFGS run reaches its
iteration cap on some datasets, so the anchor is the parameter vector at that
point rather than a converged optimum. These details can affect exact
reproduction of individual numbers and should therefore be reported with the
search protocol.

\end{document}